\documentclass[oribibl]{llncs}
\usepackage{multicol}
\usepackage{amsmath}
\usepackage{amssymb}
\usepackage{graphicx}
\usepackage{color}
\usepackage{hhline}
\usepackage{enumfrom}

\usepackage{hyperref}

\newcommand{\ignore}[1]{}

\newcommand{\mf}[1]{\mathfrak{ #1}}
\newcommand{\mbf}[1]{\mathbf{ #1}}
\newcommand{\e}{\mathbf{e}}
\newcommand{\msf}[1]{\mbox{\sf #1}}

\newcommand{\red}[1]{\textcolor{red}{#1}}

\newcommand{\bblue}[1]{#1}
\newcommand{\re}[1]{#1}
\newcommand{\bl}[1]{#1}
\newcommand{\gr}[1]{#1}

\def\Li{\mathrm{\Phi}}

\newcommand{\boxtheorem}{  \hfill $\Box$}
\newcommand{\nit}[1]{{\it #1}}

\newcommand{\mc}[1]{\mathcal{ #1}}

\newcommand{\phill}{\phantom{po}  \hfill}

\newcommand{\sfd}{{\sf d}}
\newcommand{\sfs}{{\sf s}}

\newcommand{\shap}{\mbox{\sf Shap}}
\newcommand{\shp}{\#{\it P}}
\newcommand{\es}{\mathbf{e}}
\newcommand{\resp}{\mbox{\sf Resp}}

\newcommand{\bcq}{BCQ}
\newcommand{\cq}{CQ}

\newcommand{\fo}{FO}

\title{Score-Based Explanations in Data Management and Machine Learning: An Answer-Set Programming Approach to Counterfactual Analysis}

\ignore{++
\author{{\bf Leopoldo Bertossi}$^{1,2}$ and {\bf Gabriela Reyes}$^1$}
\institute{\bf Universidad Adolfo Ib\'a\~nez$^1$\\{\bf Faculty of Engineering and Sciences}\\ and\\ Millennium Inst. for Foundational Research on Data (IMFD)$^2$\\
 Santiago, \ Chile\\ \{leopoldo.bertossi,XXX\}@uai.cl}
++}

\author{{\bf Leopoldo Bertossi}}
\institute{\bf Universidad Adolfo Ib\'a\~nez\\{\bf Faculty of Engineering and Sciences}\\ and\\ Millennium Inst. for Foundational Research on Data (IMFD)\\
 Santiago, \ Chile\\ leopoldo.bertossi@uai.cl}

\begin{document}

\thispagestyle{empty}
\pagestyle{plain}
\maketitle

\begin{abstract}
We describe some recent approaches to score-based explanations for query answers in databases and outcomes from classification models in machine learning. The focus is on work done by the author and collaborators.  Special emphasis is placed on declarative approaches based on answer-set programming to the use of counterfactual reasoning for score specification and computation. Several examples that illustrate the flexibility of these methods are shown.
\end{abstract}

\section{Introduction}

In data management and machine learning one wants {\em explanations} for certain results. For example, for query results from databases, and for outcomes of classification models in machine learning (ML). Explanations, that may come in different forms, have been the subject of philosophical enquires for a long time, but, closer to our discipline, they appear under different forms in model-based diagnosis and in causality as developed in artificial intelligence.

In the last few years, explanations that are based on {\em numerical scores} assigned to elements of a model that may contribute to an outcome have become popular. These scores attempt to capture the degree of contribution of those components to an outcome, e.g. answering questions like these: What is the contribution of this tuple to the answer to this query? \ What is the contribution of this feature value of an entity to the displayed classification of the latter?

For an example, consider a financial institution that uses a
learned classifier, $\mc{C}$, e.g. a decision tree, to determine if clients
should be granted loans or not, returning labels $0$ or $1$, resp. A
particular client, represented as an entity $\es$, applies for a loan, and the classifier
returns $\mc{C}(\es) =1$, i.e. the loan is rejected. The client requests
an explanation.

A common approach consists in giving scores to the feature values in $\es$,
to quantify their relevance in relation to the classification
outcome. The higher the score of a feature value, the more explanatory  is that value. For example, the fact that the client has value ``5"
for feature {\em Age} (in years) could have the highest score. That is, the rejection of the loan application is  due mostly to the client's very young age.

In the context of {\em explainable AI} \cite{molnar}, different scores have been proposed in the literature, and some that have a relatively older history have been applied. Among the latter we find the general {\em responsibility score} as found in {\em actual causality} \cite{HP05,CH04}. For a particular kind of application, one has to define the right  causality setting, and then apply the responsibility measure to the participating variables (see \cite{halpern15} for a newer treatment of the subject). In particular, in data management,  responsibility has been used to
quantify the strength of a tuple  as a cause for a query result \cite{suciu,tocs}.  The Shapley value, as found in {\em coalition game theory} \cite{S53}, has been used for the same purpose \cite{LBKS20}. Defining the right game function, the Shapley value assigned to a player reflects its contribution to the wealth function, which in databases corresponds to the query result.

In the context of explanations to outcomes from classification models in ML,  the Shapley value has been used to assign scores to the feature values taken by an entity that has been classified. With a particular game function, it has taken the form of the $\shap$ score, which has become quite popular and influential \cite{LetA20,LL17}.

Also recently, a {\em responsibility score}, $\resp$, has been introduced and investigated for the same purpose in \cite{deem}.  It is
based on the notions of {\em counterfactual intervention} as appearing in actual causality, and causal responsibility. More specifically,
(potential) executions of   {\em counterfactual interventions} on a {\em structural logico-probabilistic model} \ \cite{HP05} are investigated, with the purpose of answering hypothetical  questions of the form: \ {\em What would happen if we change ...?}.

Counterfactual interventions can  be used to define different forms of score-based explanations. This is the case of causal responsibility in databases (c.f. Section \ref{sec:resp}).
In {\em explainable AI}, and more commonly with classification models of ML, counterfactual interventions become hypothetical changes on the entity whose classification is being explained, to detect possible changes in the outcome  (c.f. \cite[Sec. 8]{tplp} for a more detailed discussion and references).

 Score-based explanations can also be defined in the absence of a model, and with or without {\em explicit} counterfactual interventions. Actually,
explanation scores such as $\shap$ and $\resp$ can be applied with  {\em black-box} models, in that they
use, in principle, only the input/output relation that represents the
classifier, without having access to the internal components of the
model. In this category we could find classifiers based on complex
neural networks, or XGBoost \cite{LHdR19}. They are opaque enough to
be treated as black-box models.

The $\shap$ and $\resp$ scores can also be applied with open-box models, with explicit models. Without having access to the elements of the classification model, the computation of both $\shap$ and $\resp$ is in general intractable, by their sheer definitions, and the possibly large number of counterfactual combinations that have to be considered in the computation. However, for certain classes of classifiers, e.g. decision trees, having access to the mathematical model may make the computation of $\shap$ tractable, as shown in  \cite{aaai21,guy}, where it is also shown that for other classes of explicit models, its computation  is still intractable. Something similar  applies to $\resp$ \cite{deem}.

Other explanation scores used in machine learning
appeal to the components of the mathematical model behind the
classifier. There can be all kinds of explicit models, and some are
easier to understand or interpret or use for this purpose. For
example, the {\em FICO score} proposed in \cite{CetA18}, for the FICO dataset
about loan requests, depends on the internal outputs and displayed
coefficients of two nested logistic regression models. Decision trees
\cite{mitchell}, random forests \cite{BetA84}, rule-based
classifiers, etc., could be seen as relatively easy to understand and use for providing
explanations.
\ In
\cite{deem}, the $\shap$ and $\resp$ scores were experimentally compared with each other, and also with the FICO score.

One can specify in declarative terms the counterfactual versions of tuples in databases and  of feature values in entities under classification. On this basis one can analyze diverse alternative counterfactuals, reason about them, and also specify the associated explanation scores. In these notes we do this for responsibility scores in databases and classifications models. More specifically, we use {\em answer-set programming}, a modern logic-programming paradigm that has become useful in many applications \cite{asp,gelfond}.  We show examples run with the {\em DLV} system and its extensions \cite{leone}. An important advantage of using declarative specifications resides in the possibility of adding different forms of {\em domain knowledge} and {\em semantic constraints}. Doing this with purely procedural approaches would require changing the code accordingly.

The answer-set programs (ASPs) we use are influenced by, and sometimes derived from, {\em repair programs}. These are ASPs that specify and compute the possible repairs of a database that is inconsistent with respect to a given set of integrity constraints \cite{bertossiSynth}. A useful connection between database repairs and actual causality in databases was established in \cite{tocs}. Hence, the use of repairs and repair programs.

In this article we survey some of the recent advances on the use and computation of the above mentioned  score-based explanations, both for query answering in databases and for classification in ML. This is not intended to be an exhaustive survey of the area. Instead, it is heavily influenced by our latest research. Special emphasis is placed on the use of ASPs (for many more details on this see \cite{tplp}). Taking advantage of the introduced repair programs, we also show how to specify and compute a numerical measure of inconsistency of database \cite{lpnmr19}. In this case, this would be  a {\em global} score, in contrast with the {\em local} scores applied to individual tuples in a database or feature values in an entity. \
 To introduce the concepts and techniques we will use mostly examples, trying  to convey the main intuitions and issues.

This paper is structured as follows. In Section \ref{sec:back} we provide some background material on databases and answer-set programs. In Section \ref{sec:dbs} we concentrate on causal explanations in databases, the responsibility score, and also the {\em causal-effect} score \cite{tapp16}, as an alternative to the latter. In Section \ref{sec:causASPS}, we present the causality-repair connection and repair programs for causality and responsibility  computation.  In Section \ref{sec:causAttr}, we consider causality in databases at the attribute level, as opposed to the tuple level. In Section \ref{sec:ICs}, we introduce causality and responsibility in databases that are subject to integrity constraints. In Section \ref{sec:inco} we present the global inconsistency measure for a database and the ASPs to compute it.  In Section \ref{sec:shapy}, we describe the use of the Shapley value to provide explanation scores in databases.
In Section \ref{sec:shapy}, we describe in general terms score-based explanations for classification results. In Section \ref{sec:xresp} we introduce and study the $\mbox{\sf x-Resp}$ score, a simpler version of the more general $\mbox{\sf Resp}$ score that we introduce in Section \ref{sec:resp}. In Section \ref{sec:cips} we introduce {\em counterfactual intervention programs} (CIP), which are ASPs that specify counterfactuals and the $\mbox{\sf x-Resp}$ score. In Section \ref{sec:shap}, and for completeness, we briefly present the $\mbox{\sf Shap}$ score. We end in Section \ref{sec:last} with some final conclusions.

\section{Background}\label{sec:back}

\subsection{Basics of Relational Databases}\label{sec:prel} 
  A relational schema $\mc{R}$ contains a domain of constants, $\mc{C}$,  and a set of  predicates of finite arities, $\mc{P}$. \ $\mc{R}$ gives rise to a language $\mf{L}(\mc{R})$ of first-order (FO)  predicate logic with built-in equality, $=$.  Variables are usually denoted with $x, y, z, ...$; and finite sequences thereof with $\bar{x}, ...$; and constants with $a, b, c, ...$, etc. An {\em atom} is of the form $P(t_1, \ldots, t_n)$, with $n$-ary $P \in \mc{P}$   and $t_1, \ldots, t_n$ {\em terms}, i.e. constants,  or variables.
  An atom is {\em ground} (a.k.a. a tuple) if it contains no variables. A database (instance), $D$, for $\mc{R}$ is a finite set of ground atoms; and it serves as an  \ignore{The {\em active domain} of a database $D$, denoted ${\it Adom}(D)$, is the set of constants that appear in atoms of $D$.} interpretation structure for  $\mf{L}(\mc{R})$.

A {\em conjunctive query} (\cq) is a \fo \ formula,  $\mc{Q}(\bar{x})$, of the form \ $\exists  \bar{y}\;(P_1(\bar{x}_1)\wedge \dots \wedge P_m(\bar{x}_m))$,
 with $P_i \in \mc{P}$, and (distinct) free variables $\bar{x} := (\bigcup \bar{x}_i) \smallsetminus \bar{y}$. If $\mc{Q}$ has $n$ (free) variables,  $\bar{c} \in \mc{C}^n$ \ is an {\em answer} to $\mc{Q}$ from $D$ if $D \models \mc{Q}[\bar{c}]$, i.e.  $Q[\bar{c}]$ is true in $D$  when the variables in $\bar{x}$ are componentwise replaced by the values in $\bar{c}$. $\mc{Q}(D)$ denotes the set of answers to $\mc{Q}$ from $D$. $\mc{Q}$ is a {\em Boolean conjunctive query} (\bcq) when $\bar{x}$ is empty; and when {\em true} in $D$,  $\mc{Q}(D) := \{\nit{true}\}$. Otherwise, it is {\em false}, and $\mc{Q}(D) := \emptyset$. Sometimes CQs are written  in Datalog notation as follows: \ $\mc{Q}(\bar{x}) \leftarrow P_1(\bar{x}_1),\ldots, P_m(\bar{x}_m)$.

We consider as integrity constraints (ICs), i.e. sentences of $\mf{L}(\mc{R})$: (a) {\em denial constraints} \ (DCs), i.e.  of the form $\kappa\!:  \neg \exists \bar{x}(P_1(\bar{x}_1)\wedge \dots \wedge P_m(\bar{x}_m))$,
where $P_i \in \mc{P}$, and $\bar{x} = \bigcup \bar{x}_i$; and (b) {\em functional dependencies} \ (FDs), i.e. of the form  $\varphi\!:  \neg \exists \bar{x} (P(\bar{v},\bar{y}_1,z_1) \wedge P(\bar{v},\bar{y}_2,z_2) \wedge z_1 \neq z_2)$.\footnote{The variables in $\bar{v}$ do not have to go first in the atomic formulas; what matters is keeping the correspondences between the variables in those formulas.} Here,
$\bar{x} = \bar{y}_1 \cup \bar{y}_2 \cup \bar{v} \cup \{z_1, z_2\}$, and $z_1 \neq z_2$ is an abbreviation for $\neg z_1 = z_2$. A {\em key constraint} \ (KC) is a conjunction of FDs: \  $\bigwedge_{j=1}^k \neg \exists \bar{x} (P(\bar{v},\bar{y}_1) \wedge P(\bar{v},\bar{y}_2) \wedge y_1^j \neq y_2^j)$,
with $k = |\bar{y_1}| = |\bar{y}_2|$, and generically $y^j$ stands for the $j$th variable in $\bar{y}$. For example, $\forall x \forall y \forall z(\nit{Emp}(x,y) \wedge \nit{Emp}(x,z) \rightarrow y = z)$, is an FD (and also a KC) that could say that an employee ($x$) can have at most one salary. This FD is usually written as $\nit{EmpName} \rightarrow \nit{EmpSalary}$. In the following, we will include FDs and key constraints among the DCs.

We will also consider {\em inclusion dependencies} (INDs), which are constraints of the form \ $\forall \bar{x} \exists \bar{y}(P_1(\bar{x}) \rightarrow P_2(\bar{x}^\prime,\bar{y}))$, where $P_1, P_2 \in \mc{P}$, and $\bar{x}^\prime \subseteq \bar{x}$.

\  If an instance $D$ does not satisfy the set $\Sigma$ of ICs associated to the schema, we say that $D$ is {\em inconsistent}, which is denoted with \ $D \not \models \Sigma$.

\subsection{Basics of Answer-Set Programming}\label{sec:asp}

We will give now a brief review of the basics of {\em answer-set programs} (ASPs). As customary, when we talk about ASPs, we refer to {\em disjunctive Datalog programs with weak negation and stable model semantics} \cite{GL91,gelfond}.  For this reason we will, for a given program, use the terms ``stable model" (or simply, ``model") and ``answer-set" interchangeably. \ An answer-set program $\Pi$ consists of a finite number of rules of the form

\vspace{1mm}
\begin{equation}
A_1 \vee \ldots \vee A_n \leftarrow P_1, \ldots, P_m, \nit{not} \ N_1, \ldots, \nit{not} \ N_k, \label{eq:rule}
\end{equation}

\vspace{1mm}\noindent
 where $0\leq n,m,k$, and $A_i, P_j, N_s$ are (positive) atoms, i.e. of the form $Q(\bar{t})$, where $Q$ is a predicate of a fixed arity, say, $\ell$, and $\bar{t}$ is a sequence of length $\ell$ of variables or constants.
In rule (\ref{eq:rule}), $A_1, \ldots, \nit{not} \ N_k$ are called {\em literals}, with $A_1$ {\em positive}, and $\nit{not} \  N_k$, {\em negative}. All the variables in the $A_i, N_s$ appear among those
in the $P_j$.   The left-hand side of a rule is called the {\em head}, and the right-hand side, the {\em body}.  A rule can be seen as a (partial) definition of the predicates in the head (there may be other rules with the same predicates in the head).

The constants in  program $\Pi$ form the (finite) Herbrand universe $H$ of the program. The ground version of
program $\Pi$, $\nit{gr}(\Pi)$, is obtained by instantiating the variables in $\Pi$ in all
possible ways  using
values from $H$. The Herbrand base, $\nit{H\!B}$, of $\Pi$ contains all the atoms obtained as instantiations of
predicates in $\Pi$ with constants in $H$.

A subset $M$ of $\nit{HB}$ is a model of $\Pi$ if it satisfies $\nit{gr}(\Pi)$, i.e.: For every
ground rule $A_1 \vee \ldots \vee A_n$ $\leftarrow$ $P_1, \ldots, P_m,$ $\nit{not} \ N_1, \ldots,
\nit{not} \ N_k$ of $\nit{gr}(\Pi)$, if $\{P_1, \ldots, P_m\}$ $\subseteq$ $M$ and $\{N_1, \ldots, N_k\} \cap M = \emptyset$, then
$\{A_1, \ldots, A_n\} \cap M \neq \emptyset$. $M$ is a minimal model of $\Pi$ if it is a model of $\Pi$, and $\Pi$ has no model
that is properly contained in $M$. $\nit{MM}(\Pi)$ denotes the class of minimal models of $\Pi$.
Now, for $S \subseteq \nit{HB}(\Pi)$, transform $\nit{gr}(\Pi)$ into a new, positive program $\nit{gr}(\Pi)^{\!S}$ (i.e.\  without $\nit{not}$), as follows:
Delete every rule  $A_1 \vee \ldots \vee A_n \leftarrow P_1, \ldots,P_m, \nit{not} \ N_1,$ $ \ldots,
\nit{not} \ N_k$ for which $\{N_1, \ldots, N_k\} \cap S \neq \emptyset$. Next, transform each remaining rule $A_1 \vee \ldots \vee A_n \leftarrow P_1, \ldots, P_m,$ $\nit{not} \ N_1, \ldots,
\nit{not} \ N_k$ into $A_1 \vee \ldots \vee A_n \leftarrow P_1, \ldots, P_m$. Now, $S$ is a {\em stable model} of $\Pi$ if $S \in \nit{MM}(\nit{gr}(\Pi)^{\!S})$.
Every stable model of $\Pi$ is also a minimal model of $\Pi$. Stable models are also commonly called  {\em answer sets}, and so are we going to do most of the time.

 A program is {\em unstratified} if there is a cyclic, recursive definition of a predicate that involves negation. For example, the program consisting of the rules $a \vee b \leftarrow c, \nit{not} \ d$; \ $d \leftarrow e$, and $e \leftarrow b$ is unstratified, because there is a negation in the mutually recursive definitions of $b$ and $e$. The program in Example \ref{ex:hcf} below is not unstratified, i.e. it is {\em stratified}. A good property of stratified programs is that the models can be upwardly computed following {\em strata} (layers) starting from the {\em facts}, that is from the ground instantiations of rules with empty bodies (in which case the arrow is usually omitted). We refer the reader to \cite{gelfond} for more details.

 Query answering under the ASPs comes in two forms. Under the {\em brave semantics}, a query posed to the program obtains as answers those that hold in {\em some} model of the program. However, under the {\em skeptical} (or {\em cautious}) semantics, only the answers that simultaneously hold in {\em all} the models are returned. Both are useful depending on the application at hand.

 \begin{example} \label{ex:hcf} Consider the following  program $\Pi$ that is already ground.

\vspace{-6mm}
\begin{multicols}{2}
\begin{eqnarray*}
a \vee b &\leftarrow& c\\
d &\leftarrow& b\\
a \vee b &\leftarrow& e, \ {\it not} {\it f}\\
e &\leftarrow&
\end{eqnarray*}

\phantom{o}

The program has two stable models: \ $S_1 = \{e, a\}$ and $S_2 = \{e,b,d\}$.

Each of them expresses that the atoms in it are true, and any other atom that does not belong to it, is false.
\end{multicols}

These models are incomparable under set inclusion, and are minimal models in that any proper subset of any of them is not a model of the program (i.e. does not satisfy the program). \boxtheorem
\end{example}

\section{Causal Explanations in Databases}\label{sec:dbs}

 In data management we {need to understand and compute}
{\em  why}  certain results are obtained or not, e.g. query answers,  violations of semantic conditions, etc.; and we
expect a  database system to provide {\em explanations}.

\subsection{Causal responsibility}\label{sec:causal}

Here, we will consider
{\em causality-based explanations}  \cite{suciu,suciuDEBull}, which we will illustrate by means of an example.

 \begin{example}  \label{ex:uno}  \ Consider the database ${D}$, and the Boolean conjunctive query (BCQ)

\begin{multicols}{2}

\vspace{4mm}\hspace*{1cm}\begin{tabular}{l|c|c|} \hline
$R$  & $A$ & $B$ \\\hline
 & $a$ & ${b}$\\
& $c$ & $d$\\
& ${b}$ & ${b}$\\
 \hhline{~--}
\end{tabular} \hspace*{0.5cm}\begin{tabular}{l|c|c|}\hline
$S$  & $C$  \\\hline
 & $a$ \\
& $c$ \\
& ${b}$ \\ \hhline{~-}
\end{tabular}

\phantom{oo}

\phantom{oo}

\vspace{-14mm}
\begin{equation}
\mc{Q}\!: \ \exists x \exists y ( S(x) \land R(x, y) \land S(y)). \label{eq:uno}
\end{equation}

 \noindent It holds: \ ${D \models \mc{Q}}$, i.e. the query is true in $D$.
\end{multicols}

 We ask about the  causes for $\mc{Q}$ to be true: \
A tuple ${\tau \in D}$ is
{\em counterfactual cause} for  ${\mc{Q}}$ (being true in $D$) if \ ${D\models \mc{Q}}$ \ and \ ${D\smallsetminus \{\tau\}  \not \models \mc{Q}}$.
\ In this example,   {$S(b)$ is a counterfactual cause for $\mc{Q}$}: \ If ${S(b)}$ is removed from ${D}$,
 ${\mc{Q}}$ is no longer true.

Removing a single tuple may not be enough to invalidate the query. Accordingly, a tuple ${\tau \in D}$ is  an {\em actual cause} for  ${\mc{Q}}$
if there  is a {\em contingency set} \ ${\Gamma \subseteq D}$,  such that \ ${\tau}$ \ is a   counterfactual cause for ${\mc{Q}}$ in ${D\smallsetminus \Gamma}$.
\ In this example,  ${R(a,b)}$ is an actual cause for ${\mc{Q}}$ with contingency set
${\{ R(b,b)\}}$: \ If ${R(a,b)}$ is removed from ${D}$, ${\mc{Q}}$ is still true, but further removing ${R(b,b)}$ makes ${\mc{Q}}$ false.
\boxtheorem \end{example}

Notice that every counterfactual cause is also an actual cause, with empty contingent set.   Actual causes that are not counterfactual causes need company to invalidate a query result.
 \ Now we ask  how strong are tuples as actual causes. \ To answer  this question, we appeal to the {\em responsibility} of an actual cause ${\tau}$ for ${\mc{Q}}$ \cite{suciu}, defined by:
\begin{equation*}
{\rho_{\!_D}\!(\tau) \ := \ \frac{1}{|\Gamma| \ + \ 1}},
\end{equation*}
where ${|\Gamma|}$ is the
size of a smallest contingency set, $\Gamma$, for ${\tau}$, \ and  $0$, otherwise.

\begin{example} \ (ex. \ref{ex:uno} cont.) \ The {responsibility of ${R(a,b)}$ is \  $\frac{1}{2}$} ${= \frac{1}{1 + 1}}$ \ (its several smallest contingency sets have all size ${1}$).

  ${R(b,b)}$ and ${S(a)}$ are also actual causes with responsibility  \ ${\frac{1}{2}}$; and
  ${S(b)}$ is actual (counterfactual) cause with responsibility \   $1$ ${= \frac{1}{1 + 0}}$. \boxtheorem
\end{example}

High responsibility tuples provide more interesting explanations. Causes in this case are tuples that come with their responsibilities as  ``scores".
All tuples can be seen as actual causes, but only those with non-zero responsibility score matter. \ Causality and responsibility in databases can be extended to the attribute-value level \cite{tocs,kais} (c.f.  Section \ref{sec:causAttr}).

 As we will see in Section \ref{sec:repCon}, there is a connection between database causality and  {\em repairs} of databases w.r.t. integrity constraints (ICs) \cite{bertossiSynth}. There are also  connections to {\em consistency-based diagnosis} \ and \ {\em abductive diagnosis}, that are two forms of {\em model-based diagnosis} \cite{struss}. These connections have led to new complexity and algorithmic results for causality and responsibility \cite{tocs,flairsExt}. Actually, the latter turns out to be intractable (c.f. Section \ref{sec:repCon}). In \cite{flairsExt}, causality under ICs was introduced and investigated. This allows to bring semantic and  domain knowledge into causality in databases (c.f. Section \ref{sec:ICs}).

 Model-based diagnosis is an older area of knowledge representation where explanations form the subject of investigation. In general, the diagnosis analysis is performed on a logic-based model, and certain
elements of the model are identified as explanations. Causality-based explanations are somehow more recent. In this case,
still a model is used, which is, in general, a more complex  than a database with a query.   In the case of databases, actually there is an underlying logical model,  the {\em lineage or provenance} of the query \cite{lineage,probDBs} that we will illustrate in Section \ref{sec:CE}, but it is still a relatively simple model.

\subsection{The causal-effect score}\label{sec:CE}

Sometimes, as we will see right here below, responsibility does not provide intuitive or expected results, which led to the consideration of an alternative score, the {\em causal-effect score}. We show the issues and the score by means of an example.

\begin{example} \ \label{ex:ce} Consider the database ${E}$ that represents the graph below, and the Boolean Datalog query ${\Pi}$ that is true in $E$ if there is a path from ${a}$ to ${b}$. Here, ${E \cup\Pi \models \nit{yes}}$. Tuples have global tuple identifiers (tids) in the left-most column, which is not essential, but convenient.

\begin{multicols}{3}

 \hspace*{5mm} {\footnotesize \begin{tabular}{l|c|c|} \hline
 {$E$}  &  ${A}$ &  ${B}$ \\\hline
 {$t_1$} & { $a$} &  {$b$}\\
{$t_2$}&  {$a$} &  {$c$}\\
{$t_3$}&  {$c$} &  {$b$}\\
{$t_4$}&  {$a$} &  {$d$}\\
{$t_5$}&  {$d$} &  {$e$}\\
{$t_6$}&  {$e$} &  {$b$}\\ \cline{2-3}
\end{tabular}}

 \hspace{-5mm} \includegraphics[width=3.3cm]{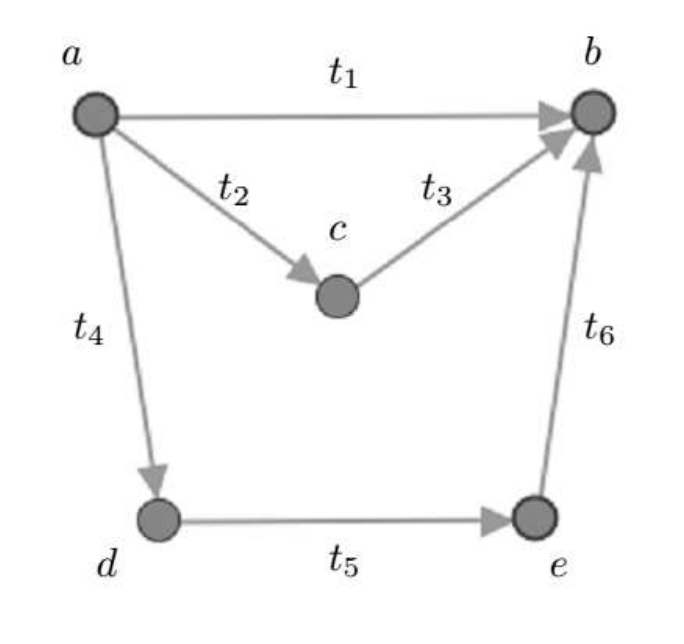}

{\small
\begin{eqnarray}
         \nit{yes }& {\leftarrow}&  {P(a,b)} \nonumber \\
        {P(x,y)}& {\leftarrow}& { E(x,y)} \nonumber \\
        {P(x,y)}& {\leftarrow}& { P(x,z), E(z,y)}\nonumber
\end{eqnarray}
}
\end{multicols}

All tuples are actual causes since every tuple appears in a path from $a$ to $b$. Also,
all the tuples  have the same causal responsibility, {$\frac{1}{3}$}, which
may be counterintuitive, considering that \ ${t_1}$ provides a direct path from ${a}$ to ${b}$.
\boxtheorem\end{example}

In \cite{tapp16}, the notion  {\em causal effect} was introduced. It is based on three main ideas, namely, the transformation, for auxiliary purposes, of the database into a probabilistic database, the expected value of a query, and
interventions on the lineage of the query. The lineage of a query represents, by means of a propositional formula, all the ways in which the query can be true in terms of the potential database tuples, and their combinations. Here, ``potential" refers to tuples that can be built with the database predicates and the database (finite) domain. These tuples may belong to the database at hand or not. For a given database, $D$, some of those atoms become true, and others false, which leads to the instantiation of the lineage (formula) o $D$.  This is all shown in the next example.

 \begin{example} \ Consider the database \ ${D}$  below, and a  BCQ. 

\begin{multicols}{2}
\hspace*{5mm}{\begin{tabular}{c|c|c|}\hline
 $R$ & $A$ & $B$ \\ \hline
  & $a$ & $b$\\
  & $a$ & $c$\\
  & $c$ & $b$\\ \hhline{~--}
  \end{tabular}~~~~~~~~~\begin{tabular}{c|c|}\hline
 $S$ & $C$ \\ \hline
  & ${b}$\\
  & $c$\\
  & \\ \hhline{~-}
  \end{tabular}}

 \noindent ${\mc{Q}: \ \exists x \exists y (R(x, y) \wedge  S(y))}$, which is true in ${D}$.
\end{multicols}

For the database $D$ in our example, the lineage of the query { instantiated on ${D}$} is given by the {propositional formula}:
\begin{equation}
{\Li_\mc{Q}(D)= (X_{R(a, b)} \wedge  X_{S(b)})  \vee (X_{R(a, c)} \wedge  X_{S(c)}) \vee (X_{R(c, b)} \wedge  X_{S(b)})}, \label{eq:lin}
\end{equation}
where   ${X_\tau}$ is a  {propositional variable} that is true iff  ${\tau \in D}$. \ Here,
  ${\Li_\mc{Q}(D)}$ \ takes value ${1}$ in ${D}$.

Now, for illustration, we want to quantify the contribution of tuple ${S(b)}$ to the query answer. \
For this purpose, we assign, uniformly and independently, probabilities to the tuples in ${D}$, obtaining a
 {\em probabilistic database} \ ${D^{{p}}}$ \cite{probDBs}.  \ Potential tuples outside ${D}$ get probability $0$.

\vspace{3mm}
   \hspace*{3cm}
   \begin{tabular}{c|c|c|c|}\hline
 $R^{{p}}$ & $A$ & $B$ & {$\mbox{prob}$}\\ \hline
  & $a$ & $b$ & \scriptsize{${\frac{1}{2}}$}\\
  & $a$ & $c$& \scriptsize{$\frac{1}{2}$}\\
  & $c$ & $b$& \scriptsize{$\frac{1}{2}$}\\ \hhline{~---}
  \end{tabular}~~~~~~~~~\begin{tabular}{c|c|c|}\hline
 $S^p$ & $C$ & $\mbox{prob}$\\ \hline
  & ${b}$& \scriptsize{${\frac{1}{2}}$}\\
  & $c$& \scriptsize{$\frac{1}{2}$}\\
  & & \\ \hhline{~--}
  \end{tabular}

 \vspace{2mm}  {The $X_\tau$'s become independent, identically distributed Boolean random variables}; \ and ${\mc{Q}}$ becomes a Boolean random variable.
Accordingly, we can ask about the probability that $\mc{Q}$ takes the truth value $1$ (or $0$) when an {\em intervention} is performed on $D$.

     Interventions are of the form ${\nit{do}(X = x)}$, meaning making ${X}$ take value ${x}$, with $x \in \{0,1\}$, in the {\em structural model}, in this case, the lineage. That is, we ask,
for \ ${\{y,x\} \subseteq \{0,1\}}$, about the conditional probability  ${P(\mc{Q} = y~|~ {\nit{do}(X_\tau = x)})}$, i.e. conditioned to  making ${X_\tau}$ false or true.

For example, with ${\nit{do}(X_{S(b)} = 0)}$ and $\nit{do}(X_{S(b)} = 1)$, the lineage in (\ref{eq:lin}) becomes, resp., and abusing the notation a bit:
\begin{eqnarray*}
\Li_\mc{Q}(D|\nit{do}(X_{S(b)} = 0) &:=&  (X_{R(a, c)} \wedge  X_{S(c)}).\\
\Li_\mc{Q}(D|\nit{do}(X_{S(b)} = 1) &:=& X_{R(a, b)}  \vee (X_{R(a, c)} \wedge  X_{S(c)}) \vee X_{R(c, b)}.
\end{eqnarray*}
On the basis of these lineages and  \ ${D^{{p}}}$, \ when ${X_{S(b)}}$ is made false, \ the probability that the instantiated lineage becomes true in {$D^p$} is:

\vspace{1mm}
  \centerline{${P(\mc{Q} = 1~|~ {\nit{do}(X_{S(b)} = 0)}) = P(X_{R(a, c)}=1) \times P(X_{S(c)}=1) = \frac{1}{4}}$.}
\vspace{1mm}
  Similarly, when ${X_{S(b)}}$ is made true, \ the probability of the lineage  becoming true in ${D^p}$ is:

\vspace{1mm}
  \centerline{${P(\mc{Q} = 1~|~{\nit{do}( X_{S(b)} = 1)}) = P(X_{R(a, b)}  \vee (X_{R(a, c)} \wedge  X_{S(c)}) \vee X_{R(c, b)} =1)}
{=  \  \frac{13}{16}}.$}

\vspace{1mm}
The {\em causal effect} of a tuple ${\tau}$ is defined by:
\begin{equation*}
{\mathcal{CE}^{D,\mc{Q}}(\tau) \ := \ \mathbb{E}(\mc{Q}~|~\nit{do}(X_\tau = 1)) - \mathbb{E}(\mc{Q}~|~\nit{do}(X_\tau = 0))}.
\end{equation*}

In particular, using the probabilities computed so far:
\begin{eqnarray*}
\mathbb{E}(\mc{Q}~|~\nit{do}(X_{S(b)} = 0)) &=&  P(\mc{Q} =1~|~\nit{do}(X_{S(b)} = 0)) \ = \  \frac{1}{4},\\
\mathbb{E}(\mc{Q}~|~\nit{do}(X_{S(b)} = 1)) &=&  P(\mc{Q} =1~|~\nit{do}(X_{S(b)} = 1)) \ = \ \frac{13}{16}.
\end{eqnarray*}

Then, \  the causal effect for the tuple ${S(b)}$ is:
   ${\mathcal{CE}^{D,\mc{Q}}(S(b)) = \frac{13}{16} - \frac{1}{4} = {\frac{9}{16}} \ > \ 0}$, showing that the tuple is relevant for the query result, with a relevance score provided by the causal effect, of   $\frac{9}{16}$.
\boxtheorem \end{example}

Let us now retake the initial example of this section.

\begin{example} \ (ex. \ref{ex:ce} cont.)  \ The Datalog query, here as a union of BCQs, has the lineage: \ ${\Li_\mc{Q}(D) = X_{t_1} \vee (X_{t_2}\wedge X_{t_3}) \vee (X_{t_4} \wedge X_{t_5} \wedge X_{t_6})}.$ It holds:
\begin{eqnarray*}
\mathcal{CE}^{D,\mc{Q}}(t_1)  &=&  {0.65625},\\
\mathcal{CE}^{D,\mc{Q}}(t_2)  &=&  \mathcal{CE}^{D,\mc{Q}}(t_3) =  0.21875,\\
\mathcal{CE}^{D,\mc{Q}}(t_4)  &=&  \mathcal{CE}^{D,\mc{Q}}(t_5) = \mathcal{CE}^{D,\mc{Q}}(t_6) = 0.09375.
\end{eqnarray*}

The causal effects are different for different tuples, and the scores are much more
intuitive than the responsibility scores.  \boxtheorem \end{example}

The definition of the causal-effect score may look  rather {\em ad hoc} and arbitrary. We will revisit it in Section \ref{sec:shapy}, where we will have yet another explanation score in databases; namely one that takes a new approach, measuring the contribution of a database tuple to a query answer through the use of the {\em Shapley value}, which  is firmly established in game theory, and is also used in several other areas \cite{S53,R88}.

The main idea is that {\em several tuples together}, much like
{players in a coalition game}, are necessary to violate an IC or produce a query result. Some may contribute more than others to the {\em  wealth distribution function} (or simply,  game function), which in this case becomes the query result, namely $1$ or $0$ if the query is Boolean, or a number if the query is an aggregation.
  The Shapley value of a tuple can be used to assign a score to its contribution. This was done in \cite{LBKS20}, and will be retaken in Section \ref{sec:shapy}. But first things first.

\section{Answer-Set Programs for Causality in Databases}\label{sec:causASPS}

In this section we will first establish a useful connection between database repairs and causes as tuples in a database. Next, we provide the basics of {\em answer-set programs} ASPs. Then, we use ASPs, taking the form
of {\em repair programs}, to specify and compute database repairs and tuples as causes for query answers. We end this section with a fully developed example using the {\em DLV} system and its extensions \cite{leone}.

\subsection{The repair connection}\label{sec:repCon}

\begin{figure}[h]
\begin{center}
\includegraphics[width=8cm]{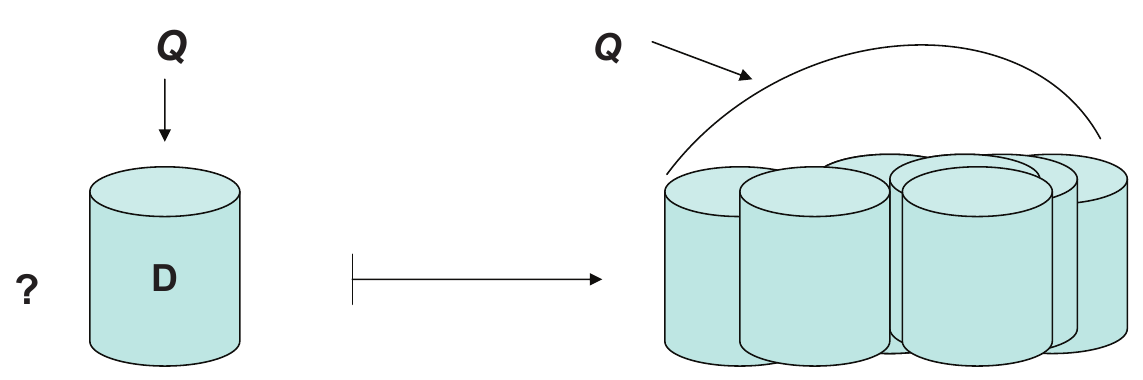}

\centerline{\hspace*{-5mm}$ \re{\not} \models \ \mbox{\bf ICs}$ \hspace{3.5cm} $\models \ \mbox{\bf ICs}$}

\label{fig:reps}
\vspace{-2mm}\caption{Database repairs and consistent query answers}
\end{center}
\end{figure}

The notion of {\em repair} of a relational database was introduced in order to formalize the notion of {\em consistent query answering} (CQA), as shown in Figure \ref{fig:reps}: If a database $D$ is inconsistent in the sense that is does not satisfy a given set of integrity constraints, $\nit{ICs}$, and a query $\mc{Q}$ is posed to $D$  (left-hand side of Figure \ref{fig:reps}), what are the meaningful, or consistent, answers to $\mc{Q}$ from $D$? They are sanctioned as those that hold (are returned as answers) from {\em all} the {\em repairs} of $D$. The repairs of $D$ are consistent instances $D'$ (over the same schema of $D$), i.e. $D' \models \nit{ICs}$, and {\em minimally depart} from $D$ \cite{pods99,bertossiSynth} (right-hand side of Figure \ref{fig:reps}).

\vspace{-6mm}

Notice that: (a) We have now a {\em possible-world} semantics for (consistent) query answering; and (b) we may use in principle any reasonable notion of distance between database instances, with each choice defining a particular {\em repair semantics}. In the rest of this section we will illustrate two classes of repairs, which have been used and investigated the most in the literature. Actually, repairs in general have got a life of their own, beyond consistent query answering.

\begin{example} \label{ex:theEx} Let us consider the following set of {\em denial constraints} (DCs) and a database $D$, whose relations (tables) are shown right here below. $D$ is inconsistent, because it violates the DCs:  it satisfies the joins that are prohibited by the DCs.

\vspace{-10mm}
\begin{multicols}{2}
\begin{eqnarray*}
\neg \exists x \exists y(P(x) \wedge Q(x,y))\\
\neg \exists x \exists y(P(x) \wedge R(x,y))
\end{eqnarray*}

\phantom{ooo}

\begin{tabular}{c|c|}\hline
$P$&A\\ \hline
&a\\
&e\\ \hhline{~-}
\end{tabular}~~~~~~~~
\begin{tabular}{c|c|c|}\hline
$Q$&A&B\\ \hline
& a & b\\ \hhline{~--}
\end{tabular}~~~~~~~~
\begin{tabular}{c|c|c|}\hline
$R$&A&C\\ \hline
& a & c\\ \hhline{~--}
\end{tabular}
\end{multicols}

We want to repair the original instance by {\em deleting tuples} from relations. Notice that, for DCs, insertions of new tuple will not restore consistency. We could change (update) attribute values though, a possibility we will consider in Section \ref{sec:causAttr}.

Here we have two {\em subset repairs}, a.k.a. {\em S-repairs}. They are subset-maximal consistent subinstances of $D$: \ $D_1 = \{P(e), Q(a,b), R(a,c)\}$ \ and \ $D_2 = \{P(e),$ $ P(a)\}$. They are consistent, subinstances of $D$, and any proper superset of them (still contained in $D$) is inconsistent. (In general, we will represent database relations as set of tuples.)

We also have {\em cardinality repairs}, a.k.a.  {\em C-repairs}. They are consistent subinstances of $D$ that minimize the {\em number} of tuples by which they differ from $D$. That is, they are maximum-cardinality consistent  subinstances. In this case, only
$D_1$ is a C-repair. \ Every C-repair is an S-repair, but not necessarily the other way around (as this example shows). \boxtheorem
\end{example}

Let us now consider a BCQ
\begin{equation}
\mc{Q}\!: \exists \bar{x}(P_1(\bar{x}_1) \wedge \cdots \wedge P_m(\bar{x}_m)),
\end{equation}
which we assume is  true in  a database $D$. \ It turns out that we can obtain the causes for $\mc{Q}$ \ to be true $D$, and their contingency sets from database repairs. In order to do this, notice that
$\neg \mc{Q}$ \ becomes a  DC \begin{equation}
\kappa(\mc{Q})\!: \   \neg \exists \bar{x}(P_1(\bar{x}_1) \wedge \cdots \wedge P_m(\bar{x}_m));
\end{equation}
and that
$\mc{Q}$ holds in $D$ \  iff \  $D$ is inconsistent w.r.t. $\kappa(\mc{Q})$.

It holds that S-repairs are associated to causes with minimal contingency sets, while C-repairs are associated to causes for $\mc{Q}$ with minimum contingency sets, and maximum responsibilities \cite{tocs}. In fact,
for a database tuple \ $\tau \in D$:
\begin{itemize}
\item [(a)] $\tau$ \ is actual cause for $\mc{Q}$ with subset-minimal contingency set $\Gamma$ \ iff \ $D \smallsetminus (\Gamma \cup \{\tau\})$ \ is an  S-repair (w.r.t. $\kappa(\mc{Q})$),
in which case, its responsibility is \ $\frac{1}{1 + |\Gamma|}$. \item[(b)]
$\tau$ \ is actual cause with minimum-cardinality contingency set $\Gamma$ \ iff \ $D \smallsetminus (\Gamma \cup \{\tau\})$ \ is C-repair,
in which case, $\tau$ is a maximum-responsibility actual cause.
\end{itemize}
Conversely, repairs can be obtained from causes and their contingency sets \cite{tocs}. These results can be extended to unions of BCQs (UBCQs), or equivalently, to sets of denial constraints.

One can exploit the connection between causes and repairs to understand the computational complexity of the former by leveraging existing results for the latter. Beyond
the fact that computing or deciding actual causes  can be done in polynomial time  in data for CQs and UCQs \cite{suciu,tocs}, one can show that
most computational problems related to responsibility are hard, because they are also hard for repairs, in particular,  for C-repairs (all this in data complexity)
\cite{lopatenko}. In particular, one can prove \cite{tocs}: (a) The {\em responsibility problem}, about deciding if a tuple has responsibility above a certain threshold, is $\nit{N\!P}$-complete for  UCQs. \
(b) Computing  $\rho_{_{\!D\!}}(\tau)$ \ is $\nit{F\!P}^{\nit{N\!P(log} (n))}$-complete for BCQs. This the {\em functional}, non-decision, version of the  responsibility problem. The complexity class involved is that of computational problems that use polynomial time with a logarithmic number of calls to an oracle in \nit{NP}. \ (c)
Deciding if a tuple $\tau$ is a most responsible cause is  $P^\nit{N\!P(log(n))}$-complete for BCQs. The complexity class is as the previous one, but for decision problems \cite{arora}.

\subsection{Answer-set programs}\label{sec:asp}

We will give now a brief review of the basics of {\em answer-set programs} (ASPs). As customary, when we talk about ASPs, we refer to {\em disjunctive Datalog programs with weak negation and stable model semantics} \cite{GL91,gelfond}.  For this reason we will, for a given program, use the terms ``stable model" (or simply, ``model") and ``answer-set" interchangeably. \ An answer-set program $\Pi$ consists of a finite number of rules of the form

\vspace{1mm}
\begin{equation}
A_1 \vee \ldots \vee A_n \leftarrow P_1, \ldots, P_m, \nit{not} \ N_1, \ldots, \nit{not} \ N_k, \label{eq:rule}
\end{equation}

\vspace{1mm}\noindent
 where $0\leq n,m,k$, and $A_i, P_j, N_s$ are (positive) atoms, i.e. of the form $Q(\bar{t})$, where $Q$ is a predicate of a fixed arity, say, $\ell$, and $\bar{t}$ is a sequence of length $\ell$ of variables or constants.
In rule (\ref{eq:rule}), $A_1, \ldots, \nit{not} \ N_k$ are called {\em literals}, with $A_1$ {\em positive}, and $\nit{not} \  N_k$, {\em negative}. All the variables in the $A_i, N_s$ appear among those
in the $P_j$.   The left-hand side of a rule is called the {\em head}, and the right-hand side, the {\em body}.  A rule can be seen as a (partial) definition of the predicates in the head (there may be other rules with the same predicates in the head).

The constants in  program $\Pi$ form the (finite) Herbrand universe $H$ of the program. The ground version of
program $\Pi$, $\nit{gr}(\Pi)$, is obtained by instantiating the variables in $\Pi$ in all
possible ways  using
values from $H$. The Herbrand base, $\nit{H\!B}$, of $\Pi$ contains all the atoms obtained as instantiations of
predicates in $\Pi$ with constants in $H$.

A subset $M$ of $\nit{HB}$ is a model of $\Pi$ if it satisfies $\nit{gr}(\Pi)$, i.e.: For every
ground rule $A_1 \vee \ldots \vee A_n$ $\leftarrow$ $P_1, \ldots, P_m,$ $\nit{not} \ N_1, \ldots,
\nit{not} \ N_k$ of $\nit{gr}(\Pi)$, if $\{P_1, \ldots, P_m\}$ $\subseteq$ $M$ and $\{N_1, \ldots, N_k\} \cap M = \emptyset$, then
$\{A_1, \ldots, A_n\} \cap M \neq \emptyset$. $M$ is a minimal model of $\Pi$ if it is a model of $\Pi$, and $\Pi$ has no model
that is properly contained in $M$. $\nit{MM}(\Pi)$ denotes the class of minimal models of $\Pi$.
Now, for $S \subseteq \nit{HB}(\Pi)$, transform $\nit{gr}(\Pi)$ into a new, positive program $\nit{gr}(\Pi)^{\!S}$ (i.e.\  without $\nit{not}$), as follows:
Delete every rule  $A_1 \vee \ldots \vee A_n \leftarrow P_1, \ldots,P_m, \nit{not} \ N_1,$ $ \ldots,
\nit{not} \ N_k$ for which $\{N_1, \ldots, N_k\} \cap S \neq \emptyset$. Next, transform each remaining rule $A_1 \vee \ldots \vee A_n \leftarrow P_1, \ldots, P_m,$ $\nit{not} \ N_1, \ldots,
\nit{not} \ N_k$ into $A_1 \vee \ldots \vee A_n \leftarrow P_1, \ldots, P_m$. Now, $S$ is a {\em stable model} of $\Pi$ if $S \in \nit{MM}(\nit{gr}(\Pi)^{\!S})$.
Every stable model of $\Pi$ is also a minimal model of $\Pi$. Stable models are also commonly called  {\em answer sets}, and so are we going to do most of the time.

 A program is {\em unstratified} if there is a cyclic, recursive definition of a predicate that involves negation. For example, the program consisting of the rules $a \vee b \leftarrow c, \nit{not} \ d$; \ $d \leftarrow e$, and $e \leftarrow b$ is unstratified, because there is a negation in the mutually recursive definitions of $b$ and $e$. The program in Example \ref{ex:hcf} below is not unstratified, i.e. it is {\em stratified}. A good property of stratified programs is that the models can be upwardly computed following {\em strata} (layers) starting from the {\em facts}, that is from the ground instantiations of rules with empty bodies (in which case the arrow is usually omitted). We refer the reader to \cite{gelfond} for more details.

 Query answering under the ASPs comes in two forms. Under the {\em brave semantics}, a query posed to the program obtains as answers those that hold in {\em some} model of the program. However, under the {\em skeptical} (or {\em cautious}) semantics, only the answers that simultaneously hold in {\em all} the models are returned. Both are useful depending on the application at hand.

 \begin{example} \label{ex:hcf} Consider the following  program $\Pi$ that is already ground.

\vspace{-6mm}
\begin{multicols}{2}
\begin{eqnarray*}
a \vee b &\leftarrow& c\\
d &\leftarrow& b\\
a \vee b &\leftarrow& e, \ {\it not} {\it f}\\
e &\leftarrow&
\end{eqnarray*}

\phantom{o}

The program has two stable models: \ $S_1 = \{e, a\}$ and $S_2 = \{e,b,d\}$.

Each of them expresses that the atoms in it are true, and any other atom that does not belong to it, is false.
\end{multicols}

These models are incomparable under set inclusion, and are minimal models in that any proper subset of any of them is not a model of the program (i.e. does not satisfy the program). \boxtheorem
\end{example}

\subsection{Repair-programs for causality in databases}\label{sec:repCau}

Answer-set programs (ASPs)  can be used to specify, compute and query S- and C-repairs. These ASPs are called ``repair programs". We will show the main ideas behind them by means of an example. For a more complete treatment see \cite{monica,bertossiSynth}.

\begin{example} \label{ex:sec} \ (example \ref{ex:uno} cont.) \ Let us consider the DC associated to the query $\mc{Q}$ in (\ref{eq:uno}):   \ $\kappa(\mc{Q}): \ \neg \exists x\exists y( S(x)\wedge R(x, y)\wedge S(y))$.

The given database is inconsistent w.r.t. $\kappa(\mc{Q})$, and we may consider its repairs.
\ Its {\em repair program}  contains the  $\bl{D}$ \ as set of  facts, now (only for convenience) with global tuple identifiers (tids) in the first attribute: \
$R(\re{1},a,b),$ $ R(2,c,d), R(3,b,b),$ $ S(4,a),$  $S(5,c), S(6,b)$.

The main rule is the properly {\em repair rule}:
\begin{equation*}
S'(\re{t_1},x,\re{\sfd}) \vee R'(t_2,x,y,\re{\sfd}) \vee S'(t_3,y,\re{\sfd}) \longleftarrow S(\re{t_1},x), R(t_2,x, y),S(t_3,y).
\end{equation*}
Here, \sfd \ is an annotation constant for ``tuple deleted".  \ This rule detects in its body (its right-hand side) a violation of the DC. If this happens, its head (its left-hand-side)
instructs the deletion of one of the tuples participating in the violation. The semantics of the program forces the choice of only one atom in the head (unless forced otherwise by other rules in the program, which does not occur in repair programs). Different choices will lead to different models of the program, and then, to different repairs.

In order to ``build" the repairs, we need the {\em collection rules}:
\begin{equation*}
S'(t,x,\re{\sfs}) \longleftarrow S(t,x), \ \nit{not} \ S'(t,x,\re{\sfd}). \ \ \ \ \mbox{ etc. }
\end{equation*}
Here,  \  \sfs \ is an annotation for ``tuple stays in repair"; and the rule collects the tuples in the original instance that have not been deleted.

There is a one-to-one correspondence between the answer-sets of the repair program and the database repairs. Actually, a  model $\bblue{M}$ of the program determines an S-repair $\bblue{D'}$ of $\bblue{D}$, as
$\re{D' \  := \ \{R(\bar{c}) \ | \ \bblue{R'(t,\bar{c},}\re{\sfs}\bblue{)} \ \in \ M\}}$. \ Conversely, every S-repair can obtained in this way.

In this example, the S-repair, \ $\bblue{D_1} = \{R(a,b), R(c,d), R(b,b), S(a), S(c)\}$, \ can be obtained from the model \
$M_1 = \{R'(\re{1},a,b,\re{\sfs}), R'(2,c,d,\sfs), R'(3,b,b,\sfs),$ $ S'(4,a,\sfs), S'(5,c,\sfs), S'(6,b,\sfd), \ldots\}$. Actually, $D_1$ is a C-repair.

There is another S-repair, $D_2 = \{\ignore{R(a,b),}R(c,d), \ignore{R(b,b),} S(a), S(c), S(b)\}$, that is associated to the model $M_2 = \{R'(\re{1},a,b,\re{\sfd}), R'(2,c,d,\sfs), R'(3,b,b,\sfd),$ $ S'(4,a,\sfs),$ $ S'(5,c,\sfs), S'(6,b,\sfs), \ldots\}$. This is not a C-repair.
\boxtheorem
\end{example}

For sets of DCs, repair programs can be made {\em normal}, i.e. non-disjunctive \cite{monica}.
CQA becomes query answering  under the {\em cautious or skeptical semantics} of ASPs (i.e. true in {\em all} repairs), which, for normal programs, is $\nit{N\!P}$-complete (in data).
This matches  the data complexity of consistent QA under DCs (c.f. \cite{bertossiSynth} for references to complexity of CQA).

Now, if we want to obtain from the program only those models that correspond to C-repairs, we can add \ {\em weak program constraints} \  (WCs), as shown in the example.

\begin{example} \label{ex:tri} (example \ref{ex:sec} cont.) \ Let us add to the program the WCs
\begin{eqnarray*}
:\sim \ R(t,\bar{x}),  R'(t,\bar{x},\sfd)\\
:\sim \ S(t,\bar{x}),  S'(t,\bar{x},\sfd).
\end{eqnarray*}
A {\em (hard) program constraint} in a program \cite{leone}, usually denoted as \begin{equation*}
:\!\!- \ P_1(\bar{x}_1), \ldots, P_1(\bar{x}_n),
\end{equation*} leads to discarding  all the models where the join in the RHS of the constraint holds.
Weak program constraints, now preceded by a ``$:\sim$", may be violated by a model, buy only the models where the {\em number} of violations of them is minimized are kept.  \ In our example, the WCs have the effect of
minimizing the number of  deleted tuples. In this way, we obtain as models only C-repairs.

In our example, we obtain C-repair $D_1$, corresponding to model $M_1$, but not S-repair $D_2$, because it is associated to model $M_2$ that is discarded due to the WCs.
\boxtheorem
\end{example}

As we already mentioned,  C-repairs are those that can be used to obtain most-responsible actual causes. Accordingly, the latter task can be accomplished through the use of repair programs with weak constraints. We illustrate this by means of our example (c.f. \cite{kais} for a detailed treatment). Actually, cause and responsibility computation  become query answering on extended repair programs.
In them, causes will be represented by means of the tids we introduced for repair programs.

\begin{example} \ (example \ref{ex:tri} cont.) \ The causes can be obtained through a new predicate, defined by the rules
\begin{eqnarray*}
\nit{Cause}(t) &\longleftarrow& R'(t,x,y,\re{\sfd}), \\
\nit{Cause}(t) &\longleftarrow& S'(t,x,\re{\sfd}),
\end{eqnarray*}
because they correspond to deleted tuples in a repair. If we want to obtain them, it is good enough to pose a query under the {\em brave semantics}, which returns what is true in {\em some} model: \
  $\Pi \ \models _\nit{brave} \nit{Cause}(t)$?

However, we would like to obtain contingency sets (for causes) and responsibilities. We will concentrate on maximum-responsibility causes and their (maximum) responsibilities, for which we assume the repair program has weak constraints, as above  (c.f. \cite{kais} for non-maximum responsibility causes).

We first introduce a new binary predicate, to collect a cause and an associated contingency tuple (which is deleted together with the tuple-cause in a same repair). This predicate is of the form \ $\nit{CauCon}(t,t')$, indicating that   $\bblue{t}$ \ is actual cause, and $\bblue{t'}$ \ is a member of the  former's contingency set. For this, for each pair of predicates $\bblue{P_i, P_j}$, not necessarily different, in the DC \ $\bblue{\kappa(\mc{Q})}$, we introduce the rule:
\begin{equation*}
\nit{CauCon}(t,t') \longleftarrow P_i'(t,\bar{x}_i,\re{\sfd}),  \ P_j'(t',\bar{x}_j,\re{\sfd}), \ t\neq t'.
\end{equation*}
This will make $t'$ a member of $t$'s contingency set.
\ In our example, \ we have the rule:
\begin{equation*}
\nit{CauCon}(t,t') \longleftarrow S'(t,x,\re{\sfd}), R'(t',u,v,\re{\sfd}),
\end{equation*}
where the inequality is not needed (for having different predicates), but also, among others,
\begin{equation*}
\nit{CauCon}(t,t') \longleftarrow S'(t,x,\re{\sfd}), S'(t',u,\re{\sfd}), t \neq t'.
\end{equation*}

In model $\bblue{M_1}$, corresponding to C-repair $\bblue{D_1}$, where there is no pair of simultaneously deleted tuples, we have no $\nit{CauCon}$ atoms.  Had model $M_2$ not been discarded due to the WCs, we would find in it (actually in its extension)  the atoms: \  $\nit{CauCon}(\re{1},\gr{3})$ \ and \ $\nit{CauCon}(\gr{3}\bblue{,}\re{1})$.
\boxtheorem
\end{example}

Contingency sets, which is what we want next, are {\em sets}, which in general are not returned as objects from an ASP. However, there are extensions of ASP and their implementations, such as {\em DLV} \cite{leone}, that, trough aggregations, support set construction. This is the case of {\em DLV-Complex} \cite{calimeri08,calimeri09}, that we have used in for running repair programs and their extensions. We do this as follows (in the program below , $t, t'$ are variables).
\begin{eqnarray}
\nit{preCon}(t,\gr{\{\}}) &\leftarrow& \nit{Cause}(t), \ \nit{not} \ \nit{Aux}_1(t) \label{eq:un}\\
\nit{Aux}_1(t) &\leftarrow& \nit{CauCon}(t,t') \label{eq:dos}\\
\nit{preCon}(t,\gr{\{t'\}}) &\leftarrow& \nit{CauCon}(t,t') \label{eq:tres}\\
\nit{preCon}(t, \gr{\nit{\#union}(C,\{t''\})}) &\leftarrow& \nit{CauCon}(t,t''), \nit{preCon}(t, C), \label{eq:cuat}\\ &&\nit{not} \ \gr{\nit{\#member}(t'',C)} \nonumber \\
\re{\nit{Con}(t, C)} &\re{\leftarrow}& \re{\nit{preCon}(t, C), \nit{not} \ \nit{Aux}_2(t,C)}  \label{eq:cuatro} \\
\nit{Aux}_2(t,C) &\leftarrow& \nit{CauCon}(t,t'), \nit{\#member}(t',C)  \nonumber
\end{eqnarray}
The auxiliary predicate in rule (\ref{eq:uno}) is used to avoid a non-safe negation. That predicate is defined by rule (\ref{eq:dos}). We are capturing here causes that do not have contingency companions, and then, they have an empty contingency set. Rule (\ref{eq:tres} is indeed redundant, but shows the main idea: a contingency companion of  a cause is taken as element into the latter's pre-contingency set.
In rule (\ref{eq:cuatro}) we have an auxiliary predicate for the same reason as in the first rule. The main idea is to stepwise keep adding  by means of set union (c.f. rule (\ref{eq:cuat}), a contingent element to a possibly partial contingency set, until there is nothing left to add. \ These maximal contingency sets are obtained with rule (\ref{eq:cuatro}).

In each model of the program with WCs, these contingency sets will have the same minimum size, and will lead to maximum responsibility causes.
Responsibility computation can be done, with numerical aggregation supported by {\em DLV-Complex},  as follows:
 \bblue{\begin{eqnarray*}
\nit{pre}\mbox{-}\nit{rho}(t,n) &\leftarrow&  \#\nit{count}\{t' : \nit{CauCon}(t,t')\} = n\\
\nit{rho}(t,m) &\leftarrow& m  * (\nit{pre}\mbox{-}\nit{rho}(t,m) +1) = 1
\end{eqnarray*}}
The first rule gives us the (minimum) size, $n$, of contingency sets, which leads to a responsibility of $\frac{1}{1 + n}$.
\ The responsibility of a (maximum responsibility) cause $\bblue{t}$ can be obtained through a query to the extended program: \    $\bblue{\Pi^e \ \models_{\nit{brave}} \nit{rho}(t,X)}?$.

 \re{ASP with WCs computation has exactly the required expressive power or computational complexity
 needed for maximum-responsibility computation  \cite{kais}.}

 \subsection{The example with {\em DLV-Complex}}\label{sec:dlv}

In this section we show in detail the running example in Section \ref{sec:repCau}, fully  specified and executed with the {\em DLV-Complex} system \cite{calimeri08,calimeri09}. C.f. \cite{kais} for more details.

\begin{example} \label{ex:rep1} (ex. \ref{ex:sec} cont.) \ The first fragment of the {\em DLV}  program below,  shows facts for  database $D$, and the disjunctive repair rule for the DC $\kappa(\mc{Q})$. In it, and in the rest of this section,  \verb+R_a, S_a, ...+ stand for $R', S', ...$ used before, with the subscript \verb+_a+ for ``auxiliary". We recall that the first attribute of a predicate holds a variable or a constant for a tid; and the last attribute of \verb+R_a+, etc. holds an annotation constant, \verb+d+ or \verb+s+, for ``deleted" (from the database) or ``stays" in a repair, resp. \ (In {\em DLV} programs, variables start with a capital letter, and constants, with lower-case.)

{\footnotesize \begin{verbatim}
    R(1,a,b). R(2,c,d). R(3,b,b). S(4,a). S(5,c). S(6,b).

    S_a(T1,X,d) v R_a(T2,X,Y,d) v S_a(T3,Y,d) :- S(T1,X),R(T2,X,Y), S(T3,Y).
    S_a(T,X,s)   :- S(T,X), not S_a(T,X,d).
    R_a(T,X,Y,s) :- R(T,X,Y), not R_a(T,X,Y,d).
\end{verbatim} }
{\em DLV} returns the stable models of the program, as follows:
{\footnotesize
\begin{verbatim}
    {S_a(6,b,d), R_a(1,a,b,s), R_a(2,c,d,s), R_a(3,b,b,s),
     S_a(4,a,s), S_a(5,c,s)}

     {R_a(1,a,b,d), R_a(3,b,b,d), R_a(2,c,d,s), S_a(4,a,s),
     S_a(5,c,s), S_a(6,b,s)}

    {S_a(4,a,d), R_a(3,b,b,d), R_a(1,a,b,s), R_a(2,c,d,s),
     S_a(5,c,s), S_a(6,b,s)}
\end{verbatim} }
These three stable models (that do not show here the original EDB) are associated to the S-repairs $D_1, D_2, D_3$, resp. Only tuples with tids $1,3,4,6$ are at some point deleted. In particular,
the first model corresponds to the C-repair \

\centerline{ $D_1 = \{R(s_4,s_3),  R(s_2,s_1), R(s_3,s_3),$ $ S(s_4), S(s_2)\}$.}

\vspace{1mm}Now, to compute causes and their accompanying deleted tuples we add  to the program the  rules defining $\nit{Cause}$ and $\nit{CauCont}$:

{\footnotesize
\begin{verbatim}
         cause(T) :- S_a(T,X,d).
         cause(T) :- R_a(T,X,Y,d).
    cauCont(T,TC) :- S_a(T,X,d), S_a(TC,U,d), T != TC.
    cauCont(T,TC) :- R_a(T,X,Y,d), R_a(TC,U,V,d), T != TC.
    cauCont(T,TC) :- S_a(T,X,d), R_a(TC,U,V,d).
    cauCont(T,TC) :- R_a(T,X,Y,d), S_a(TC,U,d).
\end{verbatim}}

Next, contingency sets can be computed by means of {\em DLV-Complex}, on the basis of the rules defining predicates $\nit{cause}$ and $\nit{cauCont}$ above:

{\footnotesize \begin{verbatim}
              preCont(T,{TC}) :- cauCont(T,TC).
    preCont(T,#union(C,{TC})) :- cauCont(T,TC), preCont(T,C),
                                 not #member(TC,C).
                  cont(T,C)   :- preCont(T,C), not HoleIn(T,C).
                  HoleIn(T,C) :- preCont(T,C), cauCont(T,TC),
                                 not #member(TC,C).
                  tmpCont(T)  :- cont(T,C), not #card(C,0).
                  cont(T,{})  :- cause(T), not tmpCont(T).
\end{verbatim}}

The last two rules  associate the empty contingency set to counterfactual causes.\ignore{
\red{This code is the direct translation of the causality rules introduced in Section \ref{sec:specTuples} as well as $\nit{tmpCont}(T)$ and $\nit{cont}(T,\{\})$ which will give us empty contingency sets for counterfactual causes, for the sake of consistency with other causes.} }

The three stable models obtained above will now be extended with $\nit{cause}$- and $\nit{cont}$-atoms, among others (unless otherwise stated, $\nit{preCont}$-, $\nit{tmpCont}$-, and $\nit{HoleIn}$-atoms will be filtered out from the output); as follows:

{\footnotesize \begin{verbatim}
    {S_a(4,a,d), R_a(3,b,b,d), R_a(1,a,b,s), R_a(2,c,d,s),
     S_a(5,c,s), S_a(6,b,s), cause(4), cause(3), cauCont(4,3),
     cauCont(3,4), cont(3,{4}), cont(4,{3})}

    {R_a(1,a,b,d), R_a(3,b,b,d), R_a(2,c,d,s), S_a(4,a,s),
     S_a(5,c,s), S_a(6,b,s), cause(1), cause(3), cauCont(1,3),
     cauCont(3,1), cont(1,{3}), cont(3,{1})}

    {S_a(6,b,d), R_a(1,a,b,s), R_a(2,c,d,s), R_a(3,b,b,s),
     S_a(4,a,s), S_a(5,c,s), cause(6), cont(6,{})}
\end{verbatim}}

The first two models above show tuple 3 as an actual cause, with one contingency set per each of the models where it appears as a cause. The last line of the third model shows that cause (with tid) 6 is the only counterfactual  cause (its contingency set is empty).

The responsibility $\rho$ can  be computed via predicate
$\nit{preRho}(T,N)$ that returns  $N = \frac{1}{\rho}$, that is the inverse of the responsibility, for each tuple with tid $T$ {\em and local to a model} that shows $T$ as a cause. We concentrate on the computation of $\nit{preRho}$ in order to compute with integer numbers, as supported by {\em DLV-Complex},\ignore{ only handles integers, thus we will work only with the integer denominator of the responsibility. Consequently, with this rule we see the tuple with the lowest pre-rho value as being the most responsible cause.} which requires setting an upper integer bound by means of \verb+maxint+, in this case, at \ignore{ must be set either in code or by using the "-N" command line option and must be at} least as large as the largest tid:

{\footnotesize \begin{verbatim}
    #maxint = 100.
    preRho(T,N + 1) :- cause(T), #int(N), #count{TC: cauCont(T,TC)} = N.
\end{verbatim}}

\noindent where the local (pre)responsibility of a cause (with tid) $T$ within a repair is obtained by counting how many instances of $\nit{cauCont}(T,?)$ exist in the model, which is the size of the local contingency set for $T$ plus 1. We obtain the following (filtered) output:

{\footnotesize \begin{verbatim}
    {S_a(4,a,d), R_a(3,b,b,d), cause(4), cause(3),
     preRho(3,2), preRho(4,2), cont(3,{4}), cont(4,{3})}

    {R_a(1,a,b,d), R_a(3,b,b,d), cause(1), cause(3),
     preRho(1,2), preRho(3,2), cont(1,{3}), cont(3,{1})}

    {S_a(6,b,d), cause(6), preRho(6,1), cont(6,{})}
\end{verbatim}}

The first model shows causes 3 and 4 with a pre-rho value of $2$. The second one, causes 3 and 1 with a pre-rho value of $2$. The last model shows cause 6 with a pre-rho value of $1$. This is also a maximum-responsibility cause, actually associated to a C-repair. Inspecting the three models, we can see that the overall pre-responsibility of cause 3 (the minimum of its pre-rho values) is $2$, similarly for cause 1. For cause 6 the overall pre-responsibility value is $1$.

Now, if we want only maximum-responsibility causes, we add weak program constraints to the program above,  to minimize the number of deletions:

{\footnotesize \begin{verbatim}
    :~ S_a(T,X,d).
    :~ R_a(T,X,Y,d).
\end{verbatim} }
\noindent {\em DLV} shows only repairs with the least number of deletions, in this case:
{\footnotesize \begin{verbatim}
    Best model: {S_a(6,b,d), R_a(1,a,b,s), R_a(2,c,d,s), R_a(3,b,b,s),
                 S_a(4,a,s), S_a(5,c,s), cause(6), preRho(6,1), cont(6,{})}
    Cost ([Weight:Level]): <[1:1]>
\end{verbatim} }

As expected, only repair $D_1$ is obtained, where only $S(6,s_3)$ is a cause, and with  responsibility $1$, making it a maximum-responsibility cause.
\boxtheorem
\end{example}

\section{Causal Explanations in Databases: Attribute-Level}\label{sec:causAttr}

In Section \ref{sec:repCon} we saw that: (a) there are different database repair-semantics; and (b) tuples as causes for query answering can be obtained from S- and C-repairs. We can extrapolate from this, and {\em define}, as opposed to only reobtain,  notions of causality on the basis of a repair semantics. This is what we will do next in order to define attribute-level causes for query answering in databases.

We may start with a repair-semantics $\mc{S}$ for databases under, say denial constraints (this is the case we need here, but we could have more general ICs). Now, we have a database
 $\bblue{D}$ and a  true BCQ $\bblue{\mc{Q}}$. As before, we have an associated (and violated) denial constraint  \ $\bblue{\kappa(\mc{Q})}$. There will be $\mc{S}$-repairs, i.e. sanctioned as such by the repair semantics $\mc{S}$. More precisely, the  repair-semantics $\mc{S}$ identifies a class \ $\bblue{\nit{Rep}^{\cal S\!}(D,\kappa(\mc{Q}))}$ \ of admissible and consistent  instances that ``minimally" depart from $\bblue{D}$. On this basis,
${\cal S}$-causes can be defined as in Section \ref{sec:repCon}(a)-(b). Of course, ``minimality" has to be defined, and comes with $\mc{S}$.

\ignore{++
 \bblue{ $t \in D$} is an \ \re{${\cal S}$-actual cause} \ for \ \bblue{ $\mc{Q}$ } \ \ iff \ \ (as on page 9 with $\mc{S}$-repairs)


 In particular, \re{prioritized repairs} \hfill {\footnotesize (Staworko et al., AMAI'12)}

 There are \re{prioritized ASPs} that can be used for repair programs \\ \phill {\footnotesize (Gebser et al., TPLP'11)}
}

We will develop this idea, at the light of an example, with a particular repair-semantics, and we will apply it to define attribute-level causes for query answering, i.e. we are interested in attribute values in tuples rather than in whole tuples. \ The repair semantics we use here is natural, but others could be used instead.

\begin{example} \label{ex:atCaus} \ Consider the database $\bblue{D}$, with tids,  and query \ $\bblue{\mc{Q}\!: \ \exists x \exists y ( S(x) \land R(x, y) \land S(y))}$, of Example \ref{ex:uno} \ and  \ the associated denial constraint \ $\kappa(\mc{Q}): \ \neg \exists x\exists y( S(x)\wedge R(x, y)\wedge S(y))$.

\vspace{-2mm}
\begin{multicols}{2}

\hspace*{1cm}\re{\begin{tabular}{l|c|c|} \hline
$R$  & A & B \\\hline
$t_1$ & $a$ & $b$\\
$t_2$& $c$ & $d$\\
$t_3$& $b$ & $b$\\
 \hhline{~--}
\end{tabular} \hspace*{0.5cm}\begin{tabular}{l|c|c|}\hline
$S$  & C  \\\hline
$t_4$ & $a$ \\
$t_5$& $c$ \\
$t_6$ & $b$\\
 \hhline{~--}
\end{tabular}  }

\noindent Since $\bblue{D \not \models \kappa(\mc{Q})}$, we need to consider repairs of $D$ w.r.t. $\kappa(\mc{Q})$.
\end{multicols}

Repairs will be obtained  by ``minimally" changing attribute  values by {\sf NULL}, as in SQL databases, which
cannot be used to satisfy a join. In this case, minimality means that {\em the set} of values changed by {\sf NULL} is minimal under set inclusion.  These are two different minimal-repairs:

\begin{multicols}{2}
\hspace*{10mm}\begin{tabular}{l|c|c|} \hline
$R$  & A & B \\\hline
$t_1$& $a$ & $b$\\
$t_2$& $c$ & $d$\\
$t_3$& $b$ & $b$\\
 \hhline{~--}
\end{tabular} \hspace*{0.5cm}\begin{tabular}{l|c|c|}\hline
$S$  & C  \\\hline
$t_4$& $a$ \\
$t_5$& $c$ \\
$t_6$& $\re{\sf NULL}$ \\ \hhline{~-}
\end{tabular}

\hspace*{10mm}\begin{tabular}{l|c|c|} \hline
$R$  & A & B \\\hline
$t_1$ & $a$ & $\re{\sf NULL}$\\
$t_2$& $c$ & $d$\\
$t_3$& $b$ & $\re{\sf NULL}$\\
 \hhline{~--}
\end{tabular} \hspace*{0.5cm}\begin{tabular}{l|c|c|}\hline
$S$  & C  \\\hline
 $t_4$& $a$ \\
$t_5$& $c$ \\
$t_6$& $b$ \\ \hhline{~-}
\end{tabular}
\end{multicols}
It is easy to check that they do not satisfy $\kappa(\mc{Q})$. \  If we denote the changed values by the tid with the position where the changed occurred, then the first repair is characterized by the set $\{t_6[1]\}$, whereas the second, by the set $\{t_1[2], t_3[2]\}$. Both are minimal since none of them is contained in the other.

Now, we could also introduce a notion of {\em cardinality-repair}, keeping those where the number of changes is a minimum. In this case, the first repair qualifies, but not the second.

These repairs identify (actually, define) the value in $\re{t_6[1]}$ as a maximum-responsibility cause for $\mc{Q}$ to be true (with responsibility $1$). Similarly,  \ $\re{t_1[2]}$ and $\re{t_3[2]}$ \ become actual causes, that do need contingent companion values,  which makes them take a responsibility of $\frac{1}{2}$ each. \boxtheorem
\end{example}

 We should emphasize that, under this semantics, we are considering attribute values participating in joins as interesting causes. A detailed treatment can be found in \cite{kais}. Of course, one could also consider as causes other attribute values in a tuple that participate in a query (being true), e.g. that in $t_3[1]$, but making them {\em non-prioritized} causes. One could also think of adjusting the responsibility measure in order to give to these causes a lower score.

\subsection{ASPs for attribute-level causality}

So as in Sections \ref{sec:repCau} and \ref{sec:dlv}, we can specify attribute-level causes via attribute-based repairs, and their ASPs. We show this at the light of an example that is given directly using {\em DLV} code (c.f. \cite{kais} for more details).

\begin{example} \label{ex:veryLast}  \ Consider the database instance
\begin{equation*}
D = \{S(a), S(b), R(b,c),R(b,d),R(b,e)\},
\end{equation*} and the BCQ $\mc{Q}\!: \exists x \exists y (S(x) \land R(x, z))$, which is true in $D$, and for which we want to find attribute-level causes.

We consider the DC corresponding to the negation of query, namely
\begin{equation*}\kappa: \ \neg \exists x \exists y (S(x) \land R(x, z)).
\end{equation*} Since $D \not \models \kappa$, $D$ is inconsistent. \
The updated instance
\begin{equation*}D_2 = \{S(a), S(\mbox{\sf NULL}), R(b,c),R(b,d),R(b,e)\}
\end{equation*} is consistent (among others obtained by updates with $\mbox{\sf NULL}$), i.e. \ $D_2 \models \kappa$.

In the {\em DLV} program below, \verb+R_a+, and \verb+S_a+ are the auxiliary predicates associated to $R$ and $S$. They accommodate annotation constants in their last argument.  The annotation constants \verb+tr+, \verb+u+, \verb+fu+ and \verb+s+ \ stand for  ``in transition" (i.e. initial or updated tuple, that could be further updated), ``has been updated", ``is final update", and ``stays in repair", resp. The tuples already contain tuple-ids. Here, \verb+T+, \verb+T2+, \verb+X+, \verb+Y+, ... are variables.

{\footnotesize \begin{verbatim}
        S(1,a).  S(2,b).  R(3,b,c).  R(4,b,d).  R(5,b,e).

         S_a(T,X,tr) :- S(T,X).
         S_a(T,X,tr) :- S_a(T,X,u).
       R_a(T,X,Y,tr) :- R(T,X,Y).
       R_a(T,X,Y,tr) :- R_a(T,X,Y,u).
\end{verbatim}}

This part of the program so far provides, as facts, the tuples in the database with their tids. It also defines each of these tuples as ``in transition". The same for those that have been updated.

The updates themselves come in the following portion of the program. In it, \verb+null+ is treated as any other constant, and can be compared with other constants (as opposed to their occurrence as $\mbox{\sf NULL}$ in SQL, where any comparison involving it is considered to be false).

The first two rules capture, in the first three atoms in the body, a violation of the constraints, i.e. a join through a non-null value, for \verb+X+. The last atom in the body of the first rule says that the value for \verb+X+
in \verb+R+ is not updated to $\nit{null}$, then, as specified in the head of the rule, it has to be updated in \verb+S+. The second rule is similar, but the other way around.\footnote{Those two {\em normal} rules could be replaced by a single disjunctive rule:\\ $S\_a(T,\nit{null},u) \vee R\_a(T,\nit{null},Y,u) \leftarrow S\_a(T,X,\nit{tr}), R\_a(T2,X,Y,\nit{tr}), X \neq \nit{null}.$ For this kind of disjunctive repair programs one can show that the normal and disjunctive versions are equivalent, i.e. they have the same models. This is because, the disjunctive program becomes {\em head-cycle free} \cite{dantsin}.}
{\footnotesize \begin{verbatim}
   S_a(T,null,u) :- S_a(T,X,tr), R_a(T2,X,Y,tr), X != null,
                    not R_a(T2,null,Y,u).
 R_a(T,null,Y,u) :- R_a(T,X,Y,tr), S_a(T2,X,tr), X != null,
                    not S_a(T2,null,u).
\end{verbatim} }
In \verb+R_a(t,m,n,fu)+ below, annotation \verb+fu+ means that the atom with tid $t$ has reached its final update (during the program evaluation). In particular, \verb+R(t,m,n)+ has already been updated, and annotation \verb+u+
should appear in the new, updated atom, say \verb+R_a(t,m1,n1,u)+, and this tuple cannot be updated any further (because relevant updateable attribute values have already been replaced by \verb+null+ \ if necessary). This is captured by the next five rules:
{\footnotesize \begin{verbatim}
     S_a(T,X,fu) :- S_a(T,X,u), not auxS1(T,X).
      auxS1(T,X) :- S(T,X), S_a(T,null,u), X != null.

  R_a(T,X,Y,fu) :- R_a(T,X,Y,u), not auxR1(T,X,Y), not auxR2(T,X,Y).
   auxR1(T,X,Y) :- R(T,X,Y), R_a(T,null,Y,u), X != null.
   auxR2(T,X,Y) :- R(T,X,Y), R_a(T,X,null,u), Y != null.
\end{verbatim} }
The final six rules collect what stays in a repair, as annotated with \verb+s+:
{\footnotesize \begin{verbatim}
     S_a(T,X,s) :- S_a(T,X,fu).
     S_a(T,X,s) :- S(T,X), not auxS(T).
        auxS(T) :- S_a(T,X,u).
   R_a(T,X,Y,s) :- R_a(T,X,Y,fu).
   R_a(T,X,Y,s) :- R(T,X,Y), not auxR(T).
        auxR(T) :- R_a(T,X,Y,u).
\end{verbatim} }
Two stable models are returned, corresponding to two attribute-based repairs: \ (we skip the atoms without annotation \verb+s+)

{\footnotesize \begin{verbatim}
{S_a(1,a,s), S_a(2,b,s), R_a(3,null,c,s), R_a(5,null,e,s), R_a(4,null,d,s)}
{S_a(1,a,s), R_a(3,b,c,s), R_a(4,b,d,s), R_a(5,b,e,s), S_a(2,null,s)}
\end{verbatim}}
The second model corresponds to the repair $D_2$ given at the beginning of this example.

We could extend the program with rules to collect the attribute values that are causes for the query to be true:
{\footnotesize \begin{verbatim}
    cause(T,1,X) :- R(T,X,Y), R_a(T,null,Z,s).
    cause(T,2,Y) :- R(T,X,Y), R_a(T,Z,null,s).
    cause(T,1,X) :- S(T,X), S_a(T,null,s).
\end{verbatim}}
Here, the second argument indicates the position where the cause, as a value, appears in a tuple. Remember that the tids are global, so having them in the first body atom in these rules will always make these rules to be evaluated with different tids, which come from the original database.

Here, we are assuming the original database does not have nulls. If it does, it is good enough to add the extra condition \verb+X != null+ in the body of the first rule, and similarly for the other rules.  Each model will return some causes. If we want them all, and we have no interest in the repairs or the complete models, we can just pose a query under the {\em brave semantics}: \ \ \verb+:- cause(U,V,W)?+ \ We will obtain all the \verb+cause+-atoms that appear in {\em some} of the models of the extended program, e.g. \verb+cause(3,1,b)+, i.e. the value \verb+b+ in the first attribute, ``\verb+1+", of tuple with id \verb+3+.
\boxtheorem
\end{example}

\section{Causes under Integrity Constraints}\label{sec:ICs}

In this section we consider tuples as causes for query answering in the more general setting where  databases are subject to integrity constraints (ICs). In this scenario, and in comparison with Section \ref{sec:causal}, not every intervention on the database is admissible, because the ICs have to be  satisfied. As a consequence, the definitions of cause and responsibility have to be modified accordingly. We illustrate the issues by means of an example. More details can be found in \cite{flairsExt,kais}.

We start assuming that a database $D$ satisfies a set of ICs, $\Sigma$, i.e. $\bblue{D \models \Sigma}$. If we concentrate on BCQs, or more, generally on monotone queries, and consider causes at the tuple level, only
instances obtained from $D$ by interventions that are tuple deletions have to be considered; and they  should  satisfy the
ICs. More precisely, for $\bblue{\tau}$ \ to be actual cause for $\bblue{\mc{Q}}$, with a contingency set $\bblue{\Gamma}$, it must hold \cite{flairsExt}:
\begin{itemize}
\item[(a)] $\re{D \smallsetminus \Gamma \ \models \ \Sigma}$, \ and \ \  $\bblue{D \smallsetminus \Gamma \ \models \ \mc{Q}}$.

\item[(b)] $\re{D \smallsetminus (\Gamma \cup \{\tau\}) \ \models \ \Sigma}$, \ and \ \  $\bblue{D \smallsetminus (\Gamma \cup \{\tau\}) \ \not \models \ \mc{Q}}$.
\end{itemize}
The {\em responsibility} of $\tau$, denoted  \bblue{$\rho_{_{\!\mc{Q}(\bar{a})\!}}^{D,\Sigma}(\tau)$}, \ is defined  as in Section \ref{sec:causal}, through minimum-size contingency sets.

\begin{example} \label{ex:ics} \ Consider the database instance $D$ as below, initially without additional ICs.
\begin{center}
{\footnotesize \begin{tabular}{c|c|c|} \hline
\nit{ Dep} & \nit{DName} &\nit{TStaff}  \\\hline
$t_1$& {\sf Computing} & \re{{\sf John}}   \\
$t_2$& {\sf Philosophy} &  {\sf Patrick}   \\
$t_3$&{\sf Math}  &  {\sf Kevin}   \\
 \hhline{~--} \end{tabular}~~~~ 
 \begin{tabular}{c|c|c|c|} \hline
\nit{Course}  & \nit{CName} & \nit{TStaff} & \nit{DName} \\\hline
$t_4$&{\sf COM08} & \re{\sf John}  & {\sf Computing} \\
$t_5$&{\sf Math01} & {\sf Kevin}  & {\sf Math} \\
$t_6$&{\sf HIST02}&  {\sf Patrick}   &{\sf Philosophy} \\
$t_7$&{\sf Math08}&  {\sf Eli}   &{\sf Math}  \\
$t_8$&{\sf COM01}&  \re{\sf John} &{\sf Computing} \\
 \hhline{~---}
\end{tabular} }
 \end{center}

Let us first consider the following open query: (The fact that it is open is not particularly relevant, because we can instantiate the query with the answer, obtaining a Boolean query.)
\begin{equation}
\bblue{\mc{Q}(\re{x})\!: \ \exists y \exists z (\nit{Dep}(y,\re{x}) \wedge
\nit{Course(z,\re{x}, y}))}.   \label{eq:A}
\end{equation}
In this case,  we get answers other that {\em yes} or {\em no}. Actually, $\bblue{\langle{\sf John}\rangle \in \mc{Q}(D)}$, the set of answers to $\mc{Q}$, and we look for causes for this particular answer. It holds: \ (a)  $\bblue{t_1}$ is a counterfactual cause; \ (b)
\gr{$t_4$ is actual cause with single minimal contingency set $\Gamma_1=\{t_8\}$}; \ (c)
$\bblue{t_8}$  is actual cause with single minimal contingency set \ $\bblue{\Gamma_2=\{t_4\}}$.

Let us now impose on $D$ the {\em inclusion dependency} (IND):
\begin{equation}
\psi: \ \ \ \bblue{\forall x \forall y \ (\nit{Dep}(x, y) \rightarrow \exists u  \  \nit{Course}(u, y, x))}, \label{eq:ind}
\end{equation}
which is satisfied by $D$.
\ Now,
 \re{$t_4$ \ $t_8$ \ are not  actual causes  anymore}; \ and  $\bblue{t_1}$ \ is still a counterfactual cause.

Let us now consider the query
\begin{equation}
\bblue{\mc{Q}_1(\re{x})\!: \ \exists y \ \nit{Dep}(y,\re{x})}. \label{eq:B}
\end{equation}
Now, $\bblue{\langle{\sf John}\rangle \in \mc{Q}_1(D)}$, and
under the IND (\ref{eq:ind}), we obtain the  \re{same causes as for \ $\bblue{Q}$}, which is not surprising considering that  \  $\re{\mc{Q} \equiv_\psi \mc{Q}_1}$, i.e. the two queries are logically equivalent under (\ref{eq:ind}).

And now, consider the query:
\begin{equation}
\bblue{\mc{Q}_2(\re{x})\!: \ \exists y \exists z \nit{Course}(z,\re{x}, y)}, \label{eq:C}
\end{equation}
for which    $\bblue{\langle{\sf John}\rangle \in \mc{Q}_2(D)}$.

For this query we consider the two scenarios, with and without imposing the IND. \ Without imposing (\ref{eq:ind}), \
$\bblue{t_4}$ and $\bblue{t_8}$ are the only actual causes, with contingency sets $\bblue{\Gamma_1 = \{t_8\}}$ and $\bblue{\Gamma_2 = \{t_4\}}$, resp.

However, imposing (\ref{eq:ind}), \  $\bblue{t_4}$ and $\bblue{t_8}$ are  still  actual causes, but we lose their smallest contingency sets
$\bblue{\Gamma_1}$ and $\bblue{\Gamma_2}$ we had before:  \  $D  \smallsetminus (\Gamma_1 \cup \{ t_4\}) \not \models \psi$, \ $D  \smallsetminus (\Gamma_2 \ \cup \ \{ t_8\}) \not \models \psi$.
\  Actually, the
smallest contingency set for $\bblue{t_4}$ \ is  \ $\bblue{\Gamma_3 = \{t_8, \re{t_1}\}}$; and for
$\bblue{t_8}$, \ $\bblue{\Gamma_4 = \{t_4, \re{t_1}\}}$.

We can see that under the IND, the responsibilities of \ $t_4$ and $t_8$ \ decrease: \  \re{$\rho_{_{\mc{Q}_2({\sf John})}}^D(t_4) = \frac{1}{2}$, but  $\rho_{_{\mc{Q}_2({\sf John})}}^{D,\psi}(t_4) =\frac{1}{3}$}. \
Tuple
 $t_1$ is not an actual cause, but it affects the responsibility of actual causes.
\boxtheorem \end{example}

Some results about causality under ICs can be obtained \cite{flairsExt}: \ (a) Causes are preserved under logical equivalence of queries under ICs, \ (b)
Without ICs, deciding causality for BCQs  is tractable, but their presence may make complexity grow. More precisely,
there are  a BCQ and an inclusion dependency
for which deciding if a tuple is an actual cause  is $\nit{N\!P}$-complete in data.

\subsection{Specifying and computing causes under integrity constraints}

 ASPs for computation of causes and responsibilities under ICs can be produced. However, Example \ref{ex:ics} shows that contingency sets may be affected by the presence of ICs.

\begin{example} \label{ex:wICS2} (ex. \ref{ex:ics} cont.) \ Database $D$ violates the DC \ $\kappa_2: \ \neg \exists z \nit{Course}(z,$ ${\sf John})$ \ associated to query $\mc{Q}_2$ and its  answer {\sf John}. Without considering $\psi$, its only minimal repair is $D' = D \smallsetminus \{\tau_4,\tau_8\}$. However, if we accept minimal repairs that also satisfy $\psi$ (when $D$ already did), then the only minimal repair is $D'' = D \smallsetminus \{\tau_1, \tau_4,\tau_8\}$. \boxtheorem
\end{example}
This example shows that, in the presence of a set of hard ICs $\Psi$, the repairs w.r.t. to another set of ICs $\Sigma$ that also satisfy $\Psi$ may not be among the repairs w.r.t. $\Sigma$ without consideration for $\Psi$. So, it is not only a matter of discarding some of the unwanted repairs w.r.t. $\Sigma$ alone.

 The example also shows that, in the presence of a hard set of ICs $\Psi$, the characterization of causes in terms of repairs (as in Section \ref{sec:causal})
 has to be revised. \ Doing this should be relatively straightforward for repairs of $D$ w.r.t.  the DCs $\Sigma$ that have origin in UBCQs, and are maximally contained in $D$ under set-inclusion, and also satisfy the hard constraints $\Psi$. Instead of giving a general approach, we show how a repair-program could be used to reobtain the results obtained in Example \ref{ex:ics}, where an inclusion dependency is our IC.

\begin{example} (exs. \ref{ex:ics} and \ref{ex:wICS2} cont.)  Without considering the IC $\psi$, the repair-program for $D$ w.r.t. the DC $\kappa_2$ is:
\begin{enumerate}
\item The extensional database as a set of facts corresponding to the table. For example, \ $\nit{Dept}(1,{\sf computing},{\sf john})$, etc.

\item Repair rule for $\kappa_2$: \ \ \ $\nit{Course}'(t,z,{\sf john},\sfd) \leftarrow \nit{Course}(t,z,{\sf john}).$

\item Persistence rule: \ \ \ $\nit{Course}'(t,x,y,\sfs) \leftarrow \nit{Course}(t,x,y), \ \nit{not} \ \nit{Course}'(t,x,y,\sfd).$
\end{enumerate}
 We have to add to this program, rules that take care of repairing w.r.t. $\psi$ in case it is violated via deletions from $\nit{Course}$:
  \begin{enumfrom}{4}
\item    $\nit{Dept}'(t',x,y,\sfd)  \leftarrow \nit{Dept}(t',x,y), \nit{not} \ \nit{aux}(y)$

\item $\nit{aux}(y) \leftarrow \nit{Course}'(t,x,y,\sfs)$.

\item $\nit{Dept}'(t,x,y,\sfs) \leftarrow \nit{Dept}(t,x,y), \ \nit{not} \ \nit{Dept}'(t,x,y,\sfd).$
\end{enumfrom}
Notice that violations of  the inclusion dependency that may arise from deletions from $\nit{Course}$ are being repaired through deletions from $\nit{Dept}$. \ The only stable model of this program corresponds to the repair in Example \ref{ex:wICS2}. \boxtheorem
\end{example}

Notice that the definition of actual cause under ICs opens the ground for a definition of a notion of {\em underlying} ({\em hidden}, {\em latent}) cause. In Example \ref{ex:ics}, $\tau_1$ could be such a cause. It is not strictly an actual cause, but it has to appear in every minimal contingency set. Similarly, Example \ref{ex:wICS2} shows that $\tau_1$ has to appear in the difference between the original instance and every minimal repair.

\section{Measuring Database Inconsistency and ASPs}\label{sec:inco}

A database $D$ is expected to satisfy a given set of integrity constraints (ICs), $\Sigma$, that come with the database schema. However, databases may be inconsistent in that those ICs are not satisfied. A natural question is: \
{\em To what extent, or how much inconsistent is \ $D$ \ w.r.t. \ $\Sigma$, in quantitative terms?}. \ This problem is about defining a {\em global numerical score} for the database, to capture its ``degree of inconsistency". This number can be interesting {\em per se}, as a measure of data quality (or a certain aspect of it), and could also be used to compare two databases (for the same schema) w.r.t. (in)consistency.

Scores for individual tuples in relation to their contribution to inconsistency can be obtain through responsibility scores for query answering, because every IC gives rise to a violation view; and a tuple contained in it can be scored. Also Shapley values can be applied (c.f. Section \ref{sec:shapy}; see also \cite{ester}).

Inconsistency measures have been introduced and investigated in knowledge representation, but mainly for  propositional theories; and, in the first-order case through grounding. In databases, it is more natural to
consider the different nature of  the combination of a database, as  a structure, and ICs, as a set of first-order formulas. It is also important to consider the asymmetry:
databases are inconsistent or not, not the combination. Furthermore, the relevant issues that are usually related to data management have to do with
algorithms  and computational complexity; actually,  in terms of the database and its size. Notice that ICs are usually few and fixed, whereas databases can be huge.

In \cite{lpnmr19}, a particular and natural  {\em inconsistency measure} (IM) was introduced and investigated. Maybe more important than the particular measure,  the research program to be developed around such an IM is particularly relevant.  More specifically, the measure was inspired by one used for functional dependencies (FDs), and reformulated and generalized {\em in terms of a class of database repairs}.
In addition to algorithms, complexity results, approximations for  hard cases of IM computation, and the dynamics of the IM under updates,  ASPs were proposed for the computation of  this measure. We concentrate on this part in the rest of this section. We use the notions and notation introduced in Section \ref{sec:repCon} and  its Example \ref{ex:theEx}.

For a database $D$ and a set of {\em denial constraints} $\Sigma$ (this is not essential, but to fix ideas), we have the classes of subset-repairs (or S-repairs), and cardinality-repairs (or C-repairs), denoted $\nit{Srep}(D,\Sigma)$ and $\nit{Crep}(D,\Sigma)$, resp. \ The following IMs are introduced:
\begin{eqnarray*}
\hspace*{1cm}\mbox{\nit{inc-deg}}^{S\!}(D,\Sigma) &:=& \frac{|D| - \nit{max}\{ |D'| ~:~D' \in \nit{Srep}(D,\Sigma)  \}}{|D|},\label{eq:s}\\
\re{\mbox{\nit{inc-deg}}^{C\!}(D,\Sigma)} &:=& \frac{|D| - \nit{max}\{ |D'| ~:~D' \in  \re{\nit{Crep}(D,\Sigma)} \}}{|D|}.\label{eq:c}
\end{eqnarray*}
 We can see that it is good enough to \re{concentrate on \ $\bl{\mbox{\nit{inc-deg}}^{C\!}(D,\Sigma)}$} since it gives the
same value as $\bl{\mbox{\nit{inc-deg}}^{S\!}(D,\Sigma)}$. \ Actually, to compute it, one C-repair is good enough. \ It is clear that
\ $\bl{0 \leq \mbox{\nit{inc-deg}}^{C\!}(D,\Sigma) \leq 1}$, \ with value $\bl{0}$ when $\bl{D}$ consistent. \ Notice that one could use other repair semantics instead of C-repairs \cite{lpnmr19}.

\begin{example} \ (example \ref{ex:theEx} cont.) \ Here, $\bl{\nit{Srep}(D,\Sigma) = \{D_1, D_2 \}}$ \ and \
$\bl{\nit{Crep}(D,\Sigma) = \{D_1 \}}$. It holds:
$\mbox{\nit{inc-deg}}^{S\!}(D,\Sigma)  = \frac{4 -|D_1|}{4} = \mbox{\nit{inc-deg}}^{C\!}(D,\Sigma) =  \frac{4 - |D_1|}{4} = \frac{1}{4}.$ \boxtheorem
\end{example}

The \re{complexity} of computing   \ $\bl{\mbox{\nit{inc-deg}}^{C\!}(D,\Sigma)}$ \ for DCs belongs to  \ $\bl{\nit{FP}^{\nit{NP(log(n))}}}$, in data complexity. Furthermore,
there is a relational schema and a set of DCs $\bl{\Sigma}$ for which
computing \ $\bl{\mbox{\nit{inc-deg}}^{C\!}(D,\Sigma)}$  is \ $\bl{\nit{FP}^{\nit{NP(log(n))}}}$-complete.
\ignore{ This result still holds for a set of two FDs of the form: \bl{$A \rightarrow B, \ B \rightarrow C$}. They fall in the class of non-simplifiable sets of FDs for which computing a  C-repair is  hard
\cite{LivshitsPODS'18}. }

It turns out that complexity and efficient computation results can be obtained via C-repairs, and we end up
confronting graph-theoretic problems. Actually, \re{C-repairs are in one-to-one correspondence with maximum-size independent sets in hypergraphs} \cite{lopatenko}.

\begin{example} \ Consider the database $\bl{D = \{A(a), B(a), C(a), D(a), E(a)\}}$, which is inconsistent w.r.t. the set of DS:
$$\bl{\Sigma= \{\neg \exists x(B(x)\wedge E(x)), \ \neg \exists x(B(x) \wedge C(x) \wedge D(x)), \ \neg \exists x(A(x) \wedge C(x))\}}.$$

We obtain the following {\em conflict hyper-graph} \ (CHG), where tuples are the nodes, and a hyperedge connects tuples that together violate a DC:\\

\begin{multicols}{2}
\includegraphics[width=3.2cm]{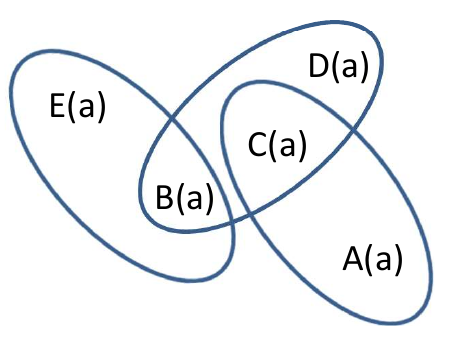}

S-repairs  are maximal \re{independent sets}: \  $\bl{D_1 = \{B(a), C(a)\}}$, \ $\bl{D_2 = \{C(a), D(a),E(a)\}}$, \ \ $\bl{D_3 = \{A(a),B(a), D(a)\}}$; and
the C-repairs are \ $\bl{D_2, \ D_3}$. \boxtheorem
\end{multicols}
\end{example}

There is a connection between C-repairs and \re{hitting-sets} (HS) of the hyperedges of the CHG: \
The removal from $\bl{D}$ of the vertices in a minimum-size HS produces a C-repair. \
The connections between hitting-sets in hypergraphs and C-repairs can be exploited for algorithmic purposes, and to obtain complexity and approximation results \cite{lpnmr19}.

It turns out that the IM can be computed via ASPs, and not surprisingly by now, via specification of  C-repairs.

\begin{example}  \ (example \ref{ex:atCaus} cont.)  Consider the  following DC and database (with tids)
\begin{eqnarray*}\kappa\!:&& \ \neg \exists x\exists y( S(x)\wedge R(x, y)\wedge S(y)),\\
D &=& \{R(1,a,b), R(2,c,d),R(3,b,b), S(4,a), S(5,c), S(6,b)\}.
\end{eqnarray*}
The repair-ASP specifying C-repairs contains the DB \ $\bl{D}$, plus the rules:
\begin{eqnarray*}
S'(t_1,x,\re{\sfd}) \vee R'(t_2,x,y,\re{\sfd}) \vee S'(t_3,y,\re{\sfd}) &\leftarrow& S(t_1,x), R(t_2,x, y),S(t_3,y),\\
S'(t,x,\re{\sfs}) &\leftarrow& S(t,x), \ \nit{not} \ S'(t,x,\re{\sfd}),\\ 
R'(t,x,y,\re{\sfs}) &\leftarrow& R(t,x,y), \ \nit{not} \ R'(t,x,y,\re{\sfd}),
\end{eqnarray*}
and \re{weak program constraints} \ (c.f. Example \ref{ex:tri}):
\begin{eqnarray*}
:\sim&&   \ R(\bar{x}),  R'(\bar{x},\sfd),\\
:\sim&& \ S(\bar{x}),  S'(\bar{x},\sfd).
\end{eqnarray*}
With them, we keep the models that  \re{minimize the number} of  deleted tuples.
\ The C-repair $\bblue{D_1}$ is represented by the model
$$M_1 = \{R'(1,a,b,\re{\sfs}), R'(2,c,d,\sfs), R'(3,b,b,\sfs), S'(4,a,\sfs), S'(5,c,\sfs), S'(6,b,\sfd), \ldots\}.$$

Now, the IM can be computed via \ $\bl{|D \smallsetminus D'|}$ \ for some (or any) C-repair  $\bl{D'}$. In this case, $D_1$.

With a  system like {\em DLV-Complex}, we can specify this set difference and compute its cardinality as a simple aggregation. More precisely, we add to the program above the rules:
\begin{eqnarray*}
\nit{Del}(t) &\leftarrow&  S'(t,x,\re{\sfd}),\\
\nit{Del}(t) &\leftarrow&  R'(t,x,y,\re{\sfd}),\\
\nit{NumDel}(n) &\leftarrow& \#count\{t : Del(t)\}=n.
\end{eqnarray*}
The first two rules collect the tids of deleted tuples. The value for $\nit{NumDel}$ defined by the third rule is the number of deleted tuples (that already takes a minimum due to the weak constraints). This number is all we need to compute
the IM. All the models, corresponding to C-repairs, will return the same number. For this reason,
there is no need to explicitly compute all stable models, their sizes,  and  compare  them.  Actually, this  value  can  be  obtained  by  means  of a query posed to the program: \  ``$:\!\!- \ \nit{NumDel}(x)?$",  that can be answered  under  the brave  semantics (returning answers that hold in some of the stable models).   In \cite[Appendix A]{lpnmr19} one can find  an extended example that uses {\em DLV-Complex} \cite{calimeri08,calimeri09} for this computation.    \boxtheorem
\end{example}

\section{The Shapley Value in Databases}\label{sec:shapy}

The Shapley value was proposed in game theory by Lloyd Shapley in 1953
\cite{S53}, to quantify the contribution of a player to a coalition game where players share a wealth function.\footnote{The original paper and related ones on the
  Shapley value can be found in the book edited by Alvin Roth
  \cite{R88}. Shapley and Roth shared the Nobel Prize in Economic
  Sciences 2012.} It has been applied in many disciplines. In particular, it has been investigated in computer science under
{\em algorithmic game theory} \cite{DBLP:books/cu/NRTV2007}, and it has been applied to many and
different computational problems. The
computation of the Shapley value is, in general, intractable. In many
scenarios where it is applied its computation turns out to be
$\shp$-hard \cite{FK92,DBLP:journals/mor/DengP94}. \ Here, the class $\shp$ contains the problems of {\em counting} the  solutions for problems in $\nit{NP}$. A typical problem in the class, actually, hard for the class, is $\#\nit{SAT}$, about counting the number of satisfying assignments for a propositional formula. Clearly, this problem cannot be easier than $\nit{SAT}$, because a solution for $\#\nit{SAT}$ immediately gives a solution for $\nit{SAT}$ \cite{arora}.

In particular, the Shapley value has been used in knowledge
representation, to measure the degree of inconsistency of a
propositional knowledge base \cite{HK10}; in machine learning to provide explanations for the outcomes of
classification models on the basis of numerical scores assigned to the participating feature values \cite{LL17} (c.f. Section \ref{sec:shap}); and in data management to
measure the contribution of a tuple to a query answer \cite{LBKS20}, which we briefly review in this section.

 Consider a set of players ${D}$,  and a
game function,  $\mc{G}:  \mc{P}(D)  \rightarrow  \mathbb{R}$, where $\mc{P}(D)$ the power set of $D$. \
 The Shapley value of player ${p}$ in ${D}$ es defined by:
  \begin{equation}{\nit{Shapley}(D,\mc{G},p):= \sum_{S\subseteq
  D \setminus \{p\}} \frac{|S|! (|D|-|S|-1)!}{|D|!}
(\mc{G}(S\cup \{p\})-\mc{G}(S))}.\label{eq:sh}\end{equation}
  Notice that here, ${|S|! (|D|-|S|-1)!}$ is the number of permutations of
${D}$ with all players  in ${S}$  coming first, then ${p}$, and then all the others. That is, this quantity
is the expected contribution of player $p$ under all possible additions of $p$ to a partial random sequence of players followed   by a random sequence of the rests of   the players. Notice the counterfactual flavor, in that there is a comparison between what happens having $p$ vs. not having it. The Shapley value is the only function that satisfies certain natural properties in relation to games. So, it is a
result of a categorical set of axioms or conditions \cite{R88}.

 Back to query answering  in databases, \ the players are tuples in the database  ${D}$. We also have a
Boolean query \ $\mc{Q}$, which becomes a   game function, as follows:  \ For \ ${S \subseteq D}$,
\begin{equation*}{\mc{Q}(S) = \left\{\begin{array}{cc} 1 & \mbox{ if } \ S \models \mc{Q},\\
0 & \mbox{ if } \ S \not \models \mc{Q}.\end{array}\right.}
\end{equation*}
With these elements we can define the Shapley value of a database tuple $\tau$:
\begin{equation*}
\nit{Shapley}(D,{\mc{Q}},{\tau}):= \sum_{S\subseteq
  D \setminus \{{\tau}\}} \frac{|S|! (|D|-|S|-1)!}{|D|!}
(\mc{Q}(S\cup \{{\tau}\})-\mc{Q}(S)).
\end{equation*}
If the query is {\em monotone}, i.e. its set of answers never shrinks when new tuples are added to the database, which is the case of conjunctive queries (CQs), among others, the difference $\mc{Q}(S\cup \{{\tau}\})-\mc{Q}(S)$ is always $1$ or $0$, and the average in the definition
of the Shapley value returns a value between $0$ and $1$. \
This value quantifies the contribution of tuple \ ${\tau}$ \ to the query result. It was introduced and investigated in \cite{LBKS20},
for BCQs and some aggregate queries defined over CQs. We report on some of the findings in the rest of this section. The analysis has been extended to queries with  negated atoms in CQs \cite{ester2}.

A main result obtained in \cite{LBKS20} is about the complexity of computing this Shapley score. The following
{\em Dichotomy Theorem} holds:  \ For ${\mc{Q}}$ a BCQ without self-joins, if \ ${\mc{Q}}$ \ is {\em hierarchical}, then ${\nit{Shapley}(D,\mc{Q},\tau)}$ can be computed in polynomial-time
(in the size of $D$); otherwise, the problem is \ {$\nit{\#P}$-complete}.

Here,  ${\mc{Q}}$ \ is {hierarchical} if for every two existential variables ${x}$ and ${y}$, it holds: \ (a)
{$\nit{Atoms}(x) \subseteq \nit{Atoms}(y)$}, \ or
 {$\nit{Atoms}(y) \subseteq \nit{Atoms}(x)$}, \ or
 {$\nit{Atoms}(x) \cap \nit{Atoms}(y) = \emptyset$}.
 \ For example,  ${\mc{Q}: \ \exists x \exists y \exists z(R(x,y) \wedge S(x,z))}$, for which \
{$\nit{Atoms}(x)$ $ = \{R(x,y),$ $ \ S(x,z)\}$,  \ $\nit{Atoms}(y) = \{R(x,y)\}$,  \ $\nit{Atoms}(z) = \{S(x,z)\}$}, is
hierarchical. \
However, \ ${\mc{Q}^{\nit{nh}}: \ \exists x \exists y({R(x) \wedge S(x,y) \wedge T(y)})}$, for which \
  {$\nit{Atoms}(x) = \{R(x), \ S(x,y)\}$,  \ $\nit{Atoms}(y) = \{S(x,y), T(y)\}$}, is not hierarchical.

These are the same criteria for (in)tractability that apply to evaluation of BCQs  over probabilistic databases \cite{probDBs}. However, the same proofs do not  apply, at least not straightforwardly.
The intractability result uses query  ${\mc{Q}^{\nit{nh}}}$ \ above, and a
reduction from {counting independent sets in a bipartite graph}.

The {dichotomy results can be extended  to summation} over CQs, with the same conditions and cases. This is because the
Shapley value, as an expectation, is linear. \
{Hardness extends to aggregates {\sf max}, {\sf min}, and {\sf avg} over non-hierarchical queries}.

For the hard cases, there is, as established in \cite{LBKS20},  an {\em approximation result}: \ For every fixed BCQ $\mc{Q}$ (or summation over a  CQ), there is a {\em multiplicative fully-polynomial randomized approximation scheme} (FPRAS) \cite{arora}, $A$, with
$${P(\tau \in D ~|~ \frac{\nit{Shapley}(D,\mc{Q},\tau)}{1+\epsilon} \leq A(\tau,\epsilon,\delta) \leq (1 + \epsilon)\nit{Shapley}(D,\mc{Q},\tau)\}) \geq 1 - \delta}.$$

 A related and popular score, in coalition games and other areas, is the  {\em Bahnzhaf Power Index}, which is similar to the Shapley value, but the order of players is ignored, by considering subsets of players rather than permutations thereof. It is defined by:
\begin{equation*}{\nit{Banzhaf}(D,\mc{Q},\tau) := \frac{1}{2^{|D|-1}} \cdot \sum_{S \subseteq (D\setminus \{\tau\})} (\mc{Q}(S \cup \{\tau\}) - \mc{Q}(S))}.
\end{equation*}
The Bahnzhaf-index is  also difficult to compute; provably \#{\it P}-hard in general. The results in \cite{LBKS20} carry over to this index when applied to query answering.
\ In \cite{LBKS20} it was proved that the causal-effect score of Section \ref{sec:CE} coincides with the Banzhaf-index, which gives to the former an additional justification.

\section{Score-Based Explanations for Classification}\label{sec:cla}

\begin{figure}[h]
\centerline{\includegraphics[width=5cm]{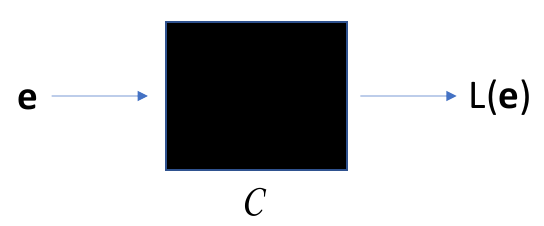}}
\vspace{-3mm}\caption{A black-box classifier} \label{fig:bb}
\end{figure}

Let us consider, as in Figure \ref{fig:bb}, a classifier, $\mc{C}$, that receives as input the representation of a entity, $\e =\langle x_1, \ldots, x_n\rangle$, as a record of feature values, and returns as an output a label, $L(\e)$, corresponding to the classification of input $\e$.
In principle, we could see $\mc{C}$ as a black-box, in the sense that only by direct interaction with it,  we have access to its input/output relation. That is, we may have no access to the mathematical classification model inside $\mc{C}$.

\vspace{-5mm}To simplify the presentation, we will assume that the  classifier is {\em binary}, that is, for every entity $\e$, $L(\e)$ takes one of two possible values, e.g. in $\{0,1\}$. \ For example, a client of a financial institution requests a loan, but the classifier, on the basis of his/her feature values (e.g. for $\mbox{\sf EdLevel}$, $\mbox{\sf Income}$, $\mbox{\sf Age}$, etc.)  assigns the label $1$, for rejection. An explanation may be  requested by the client, independently from the kind of classifier that is being used. The latter could be an explicit classification model, e.g. a classification tree or \ a {logistic regression model}. In these cases, we might be in a better position to given an explanation, because we can inspect the internals of the model \cite{rudin}. However, we will put ourselves in the ``worst scenario" in which we do not have access to the internal model. That is, we are confronted to a black-box classifier.

An approach to explanations that has become popular, specially in the absence of the model, assigns numerical {\em scores} to the feature values for an entity, trying to answer the question about which   of the feature  values contribute the most to the received label.

\begin{example} \label{ex:tree} Reusing  a popular example from \cite{mitchell}, let us consider the set of features $\mc{F} = \{\mathsf{Outlook}, \mathsf{Humidity}, \mathsf{Wind}\}$, with $\nit{Dom}(\mathsf{Outlook}) =$ $\{\mathsf{sunny}, \mathsf{overcast},$
$\mathsf{rain}\}$,  $\nit{Dom}(\mathsf{Humidity}) =$ $\{\mathsf{high}, \mathsf{normal}\}$, $\nit{Dom}(\mathsf{Wind}) =$ $\{\mathsf{strong},$ $ \mathsf{weak}\}$. An entity under classification has a value for each of the features, e.g. $\e = \nit{ent}(\mathsf{sunny},$ $ \mathsf{normal}, \mathsf{weak})$, and represents a particular weather condition. The problem consists in deciding about playing tennis or not under the conditions represented by that entity, which can be captured as a classification problem, with labels ``$\mathsf{yes}$" or ``$\mathsf{no}$". \vspace{5mm}

\begin{center}\begin{tabular*}{7cm}{|c|}\cline{1-1}\\
\phantom{o} \hspace{6cm} \phantom{o}\\
\phantom{o}\\
\phantom{o}\\
\phantom{o}\\
\phantom{o}\\
\phantom{o}\\
\phantom{o}\\
\phantom{o}\\
\phantom{o}\\
\cline{1-1}
\end{tabular*}
\end{center}

\vspace{-5cm}
\begin{figure}[h]
\begin{center}
\includegraphics[width=5.8cm]{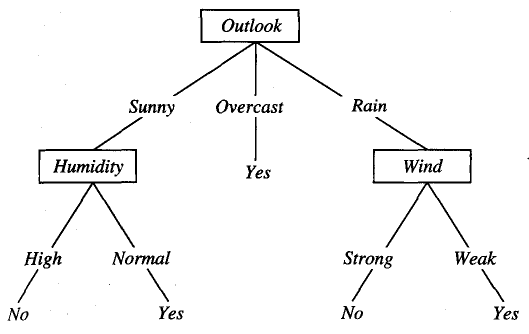}
\caption{A Decision Tree}\label{fig:tree}
\end{center}
\end{figure}

\vspace{-3mm}In this case, the binary classifier is given as a decision-tree, as shown in Figure \ref{fig:tree}. It could be displayed by double-clicking on the black box in Figure \ref{fig:bb}. The decision is computed by following the feature values along the branches of the tree.  The entity $\e$ at hand gets label $\msf{yes}$.
\boxtheorem
\end{example}

Score-based methodologies are sometimes based on {\em counterfactual interventions}: What would happen with the label if we change this particular value, leaving the others fixed? Or the other way around: What if we leave this value fixed, and change the others? The resulting labels from these counterfactual interventions can be aggregated in different ways, leading to a score for the feature value under inspection.

A be more concrete, we can use the previous example, to detect and quantify the relevance (technically, the responsibility) of a feature value in $\e = \nit{ent}(\mathsf{sunny}, \underline{\mathsf{normal}}, \mathsf{weak})$, say for feature $\msf{Humidity}$ (underlined), by {\em hypothetically intervening} its value. In this case, if we change it from $\msf{normal}$ to $\msf{high}$, we obtain a new entity $\e' = \nit{ent}(\mathsf{sunny}, \underline{\mathsf{high}}, \mathsf{weak})$, a {\em counterfactual version} of $\e$. If we input this entity into the classifier, we now obtain the label $\msf{no}$. This is an indication that the original feature value for $\msf{Humidity}$ is indeed relevant for the original classification.

In the next two sections we briefly introduce two scores. Both can be applied with open-box or black-box models. \
In both cases, we consider a finite set of features $\mc{F}$, with each feature $F \in \mc{F}$ having a finite domain, $\nit{Dom}(F)$, where $F$, as function, takes its values. The features are applied to entities $\e$ in a population $\mc{E}$ of them. Actually, we identify the entity $\e$ with the record (or tuple) formed by the values the features take on it: \ $\e = \langle F_1(\e), \ldots, F_n(\e)\rangle$.
\ Now, entities in $\mc{E}$ go through a binary {\em classifier}, $C$, that returns {\em labels} for them. We will assume the  labels are \ $1$ or $0$. For example, the bank could have a classifier that automatically decides, for an entity, if it is worthy of a loan ($0$) or not ($1$).

\section{The $\mbox{\sf x-Resp}$ Score}\label{sec:xresp}

Assume that an entity $\e$ has received the label $1$ by the classifier $\mc{C}$, and we want to explain this outcome by assigning
numerical scores to $\e$'s feature values, in such a way, that a higher score for a feature value reflects that it has been important for the outcome. We do this now using the  $\mbox{\sf x-Resp}$ score, whose definition we illustrate by means of an example (c.f. \cite{rr20,tplp} for detailed treatments). For simplicity and for the moment, we will assume the features are also binary, i.e. they propositional, taking the values $\mbox{\sf true}$ or    $\mbox{\sf false}$ (or $1$ and $0$, resp.) In Section \ref{sec:resp}, we consider a more general case.

\vspace{-1cm}
\begin{figure}[h]
\begin{center}
\includegraphics[width=11cm]{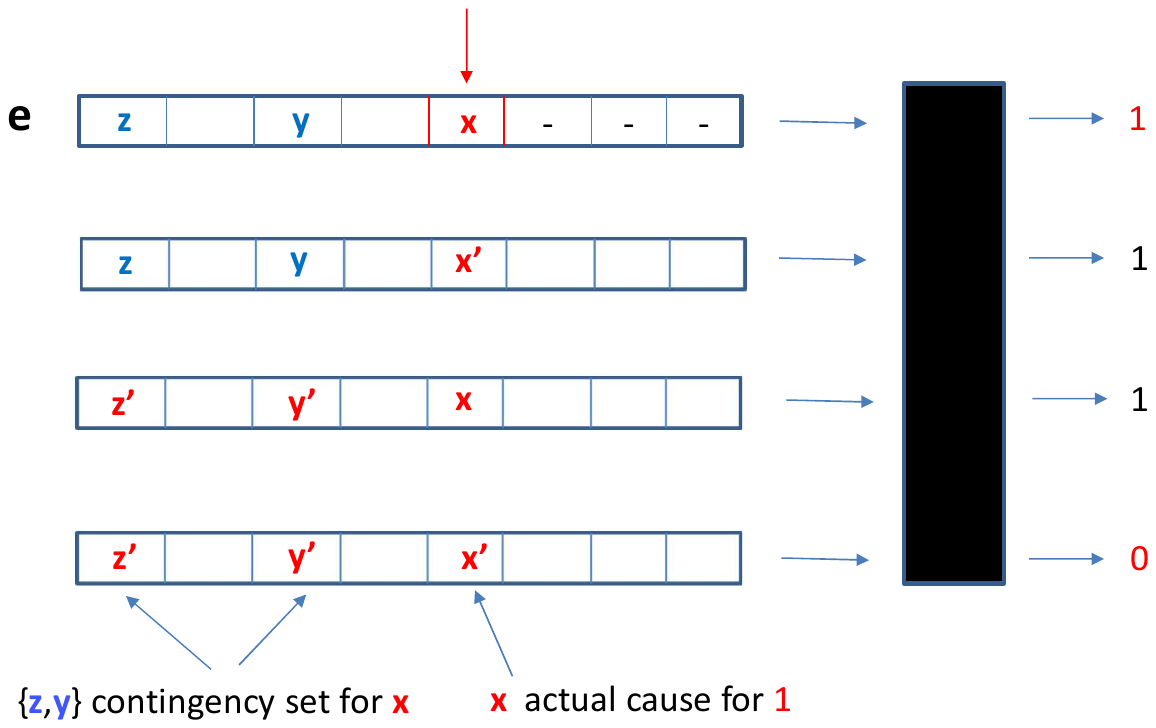}

\vspace{-2.8cm}\caption{Classified entity and its counterfactual versions}\label{fig:cl}
\end{center}
\end{figure}

\vspace{-8mm} \begin{example} In Figure \ref{fig:cl}, the black box is the classifier $\mc{C}$. An entity $\e$ has gone through it obtaining label $1$, shown in the first row in the figure. We want to assign a score to the feature value $\mathbf{x}$ for a feature $F \in \mc{F}$. \ We proceed, counterfactually, changing the value   $\mathbf{x}$ into $\mathbf{x}'$, obtaining a counterfactual version $\e_1$ of $\e$. We classify $\e_1$, and we still get the outcome $1$ (second row in the figure). In between, we may counterfactually change other feature values, $\mathbf{y}, \mathbf{z}$ in $\e$, into $\mathbf{y}', \mathbf{z}'$, but keeping $\mathbf{x}$, obtaining entity $\e_2$, and the outcome does not change (third row). However, if we change in $\e_2$, $\mathbf{x}$ into $\mathbf{x}'$, the outcome does change (fourth row).

 This shows that the value $\mathbf{x}$ is relevant for the original output, but, for this outcome,  it needs company, say of the feature values $\mathbf{y}, \mathbf{z}$ in $\e$. Proceeding as in actual causality as applied to tuples in a database in relation to query answering (c.f. Section \ref{sec:causal}), we can say that the feature value $\mathbf{x}$ in $\e$ is an {\em actual cause} for the classification, that needs a {\em contingency set} formed by the values $\mathbf{y}, \mathbf{z}$ in $\e$. In this case, the contingency set has size $2$. If we found a contingency set for $\mathbf{x}$ of size $1$ in $\e$, we would consider $\mathbf{x}$ even more relevant for the output.
\boxtheorem
 \end{example}

On this basis, we can define \cite{rr20,tplp}: \ (a) ${\mathbf{x}}$ \ is a {\em counterfactual explanation} for \ ${L(\mathbf{e}) =1}$ \ if \ ${L(\mathbf{e}\frac{{\mathbf{x}}}{\mathbf{x}^\prime}) = 0}$, \ for some ${\mathbf{x}^\prime \in \nit{Dom}(F)}$ (the domain of feature $F$). (Here we use the common notation $\mathbf{e}\frac{{\mathbf{x}}}{\mathbf{x}^\prime}$ for the entity obtained by replacing $\mathbf{x}$ by $\mathbf{x}'$ in $\mathbf{e}$.) \ (b)
 ${\mathbf{x}}$ \ is an {\em actual explanation} for \ ${L(\mathbf{e}) =1}$ \ if there is a set  of values ${\mathbf{Y}}$ in ${\mathbf{e}}$, with ${{\mathbf{x}} \notin \mathbf{Y}}$, and new values ${\mathbf{Y}^\prime \cup \{\mathbf{x}^\prime\}}$, such that \ ${L(\mathbf{e}\frac{\mathbf{Y}}{\mathbf{Y}^\prime}) = 1}$  and \ ${L(\mathbf{e}\frac{{\mathbf{x}}\mathbf{Y}}{\ \mathbf{x}^\prime\mathbf{Y}^\prime}) = 0}$.

Contingency sets may come in sizes from $0$ to $n-1$ for feature values in records of length $n$. Accordingly, we can define for the actual cause $\mbf{x}$: If ${\mbf{Y}}$ is a minimum-size contingency set for $\mbf{x}$, $\mbox{\sf x-Resp}(\mbf{x}) := \frac{1}{1 + |\mbf{Y}|}$; \ and as $0$ when $\mbf{x}$ is not an actual cause.

{\em We will reserve the notion of {\em counterfactual explanation} for (or {\em counterfactual version} of) an input entity $\e$ for any entity $\e'$ obtained from $\e$ by modifying feature values in $\e$ and that leads to a different label, i.e. $L(\e) \neq L(\e')$. Notice that from such an $\e'$ we can read off actual causes for $L(\e)$ as feature values, and contingency sets for those actual causes. It suffices to compare $\e$ with $\e'$.}

In Section \ref{sec:cips} we give a detailed example that illustrates these notions, and also show the use of ASPs for the specification and computation of counterfactual versions of a given entity, and the latter's $\mbox{\sf x-Resp}$ score.

\section{Counterfactual-Intervention Programs}\label{sec:cips}

Together with illustrating the notions introduced in Section \ref{sec:xresp}, we will introduce, by means of an example,  {\em Counterfactual Intervention Programs} (CIPs). They are ASPs that specify the counterfactual versions of a given entity, and also, if so desired, only the {\em maximum-responsibility} counterfactual explanations, i.e. counterfactual versions that lead to a maximum $\mbox{\sf x-Resp}$ score. See \cite{tplp} for many more details and examples.

\begin{example} \label{ex:tree2} (example \ref{ex:tree} continued) \ We present now the CIP for the classifier based on the decision-tree, in {\em DLV-Complex} notation. We use annotation constants \verb+o+, for ``original entity", \verb+do+, for ``do a counterfactual intervention" (a single change of feature value), \verb+tr+, for ``entity in transition", and \verb+s+, for ``stop, the label has changed". We explain the program as we present it, and also by inserting comments in the {\em DLV} code.

Notice that after the facts, that include the domains and the input entity,  we find the rule-based specification of the decision tree. The \verb+ent+ predicate, for ``entity", uses an entity identifier (eid) in its first argument.

{\footnotesize
\begin{verbatim}
 % facts:
    dom1(sunny). dom1(overcast). dom1(rain). dom2(high). dom2(normal).
    dom3(strong). dom3(weak).
    ent(e,sunny,normal,weak,o).   % original entity at hand

 % specification of the decision-tree classifier:
    cls(X,Y,Z,1) :- Y = normal, X = sunny, dom1(X), dom3(Z).
    cls(X,Y,Z,1) :- X = overcast, dom2(Y), dom3(Z).
    cls(X,Y,Z,1) :- Z = weak, X = rain, dom2(Y).
    cls(X,Y,Z,0) :- dom1(X), dom2(Y), dom3(Z), not cls(X,Y,Z,1).

 % transition rules: the initial entity or one affected by a value change
    ent(E,X,Y,Z,tr) :- ent(E,X,Y,Z,o).
    ent(E,X,Y,Z,tr) :- ent(E,X,Y,Z,do).

 % counterfactual rule: alternative single-value changes
    ent(E,Xp,Y,Z,do) v ent(E,X,Yp,Z,do) v ent(E,X,Y,Zp,do) :-
                       ent(E,X,Y,Z,tr), cls(X,Y,Z,1), dom1(Xp), dom2(Yp),
                       dom3(Zp), X != Xp, Y != Yp, Z!= Zp,
                       chosen1(X,Y,Z,Xp), chosen2(X,Y,Z,Yp),
                       chosen3(X,Y,Z,Zp).
 \end{verbatim} }
In this rule's body we find the ``choice operator". It is a predicate (to de defined next in the program), say  $\nit{chosen}_1(x,y,z,x')$, that, for each combination of values $(x,y,z)$ ``chooses" a single value for $x'$. This new value can be used to replace a value in the first argument of the entity.  Similarly for $\nit{chosen}_2(x,y,z,y')$ and $\nit{chosen}_3(x,y,z,z')$. They can be defined by means of the next rules in the program \cite{zaniolo}.

{\footnotesize
\begin{verbatim}
 % definitions of "chosen" predicates:
    chosen1(X,Y,Z,U) :- ent(E,X,Y,Z,tr), cls(X,Y,Z,1), dom1(U), U != X,
                        not diffchoice1(X,Y,Z,U).
    diffchoice1(X,Y,Z, U) :- chosen1(X,Y,Z, Up), U != Up, dom1(U).
    chosen2(X,Y,Z,U) :- ent(E,X,Y,Z,tr), cls(X,Y,Z,1), dom2(U), U != Y,
                        not diffchoice2(X,Y,Z,U).
    diffchoice2(X,Y,Z, U) :- chosen2(X,Y,Z, Up), U != Up, dom2(U).
    chosen3(X,Y,Z,U) :- ent(E,X,Y,Z,tr), cls(X,Y,Z,1), dom3(U), U != Z,
                        not diffchoice3(X,Y,Z,U).
    diffchoice3(X,Y,Z, U) :- chosen3(X,Y,Z, Up), U != Up, dom3(U).

 % Not going back to initial entity (program constraint):
    :- ent(E,X,Y,Z,do), ent(E,X,Y,Z,o).
  \end{verbatim} }
The last rule is a (hard) {\em program constraint} that avoids going back to the initial entity by performing value changes. This constraint makes the ASP evaluation engine discard those models where this happen \cite{leone}.

{\footnotesize
\begin{verbatim}
 % stop when label has been changed:
    ent(E,X,Y,Z,s) :- ent(E,X,Y,Z,do), cls(X,Y,Z,0).

 % collecting changed values for each feature:
    expl(E,outlook,X)   :- ent(E,X,Y,Z,o), ent(E,Xp,Yp,Zp,s), X != Xp.
    expl(E,humidity,Y)  :- ent(E,X,Y,Z,o), ent(E,Xp,Yp,Zp,s), Y != Yp.
    expl(E,wind,Z)      :- ent(E,X,Y,Z,o), ent(E,Xp,Yp,Zp,s), Z != Zp.

    entAux(E) :- ent(E,X,Y,Z,s).        % auxiliary predicate to
                                        % avoid unsafe negation
                                        % in the constraint below
    :- ent(E,X,Y,Z,o), not entAux(E).   % discard models where
                                        % label does not change

 % computing the inverse of x-Resp:
    invResp(E,M) :- #count{I: expl(E,I,_)} = M, #int(M), E = e.
\end{verbatim} }

The last rule returns, for a given entity, the number of values that have been changed in order to reach a counterfactual version of that entity. The inverse of this value can be used to compute a $\mbox{\sf x-Resp}$ score (the $\frac{1}{1+|\mathbf{Y}|}$ in Section \ref{sec:resp}).

 Two counterfactual versions of $\e$ are obtained, as represented by the two essentially different stable models of the program, and determined by the atoms with the  annotation \verb+s+ (below, we keep in them  only the most relevant atoms, omitting initial facts and choice-related atoms):

 {\footnotesize
\begin{verbatim}
 {ent(e,sunny,normal,weak,o), cls(sunny,normal,strong,1),
  cls(sunny,normal,weak,1), cls(overcast,high,strong,1),
  cls(overcast,high,weak,1), cls(rain,high,weak,1),
  cls(overcast,normal,weak,1), cls(rain,normal,weak,1),
  cls(overcast,normal,strong,1), cls(sunny,high,strong,0),
  cls(sunny,high,weak,0), cls(rain,high,strong,0),
  cls(rain,normal,strong,0), ent(e,sunny,high,weak,do),
  ent(e,sunny,high,weak,tr), ent(e,sunny,high,weak,s),
  expl(e,humidity,normal),invResp(e,1)}

 {ent(e,sunny,normal,weak,o), cls(sunny,normal,strong,1),...,
  cls(rain,normal,strong,0), ent(e,rain,normal,strong,do),
  ent(e,rain,normal,strong,tr), ent(e,rain,normal,strong,s),
  expl(e,outlook,sunny), expl(e,wind,weak), invResp(e,2)}
\end{verbatim} }

The first model shows the classifiers as a set of atoms, and, in  its second last line, that \verb+ent(e,sunny,high,weak,s)+ is a counterfactual version (with label $0$) of the original entity $\e$, and is obtained from
the latter by means of changes of values in feature $\mathsf{Humidity}$, leading to an inverse score of $1$. \ The second model shows a different counterfactual version of $\e$, namely \verb+ent(e,rain,normal,strong,s)+, now obtained by changing values for features $\mathsf{Outlook}$ and $\mathsf{Wind}$, leading to an inverse score of $2$.

Let us now add, at the end of the program the following weak constraints:

{\footnotesize \begin{verbatim}
 % Weak constraints to minimize number of changes:         (*)
    :~ ent(E,X,Y,Z,o), ent(E,Xp,Yp,Zp,s), X != Xp.
    :~ ent(E,X,Y,Z,o), ent(E,Xp,Yp,Zp,s), Y != Yp.
    :~ ent(E,X,Y,Z,o), ent(E,Xp,Yp,Zp,s), Z != Zp.
\end{verbatim} }
If we run the program with them, the number of changes is minimized, and we basically obtain only the first model above, corresponding to the counterfactual entity
$\e' = \nit{ent}(\mathsf{sunny},\mathsf{high},\mathsf{weak})$. This is a maximum-responsibility counterfactual explanation. \boxtheorem
\end{example}
As can be seen at the light of this example, more complex rule-based classifiers could be defined inside  a CIP. It is also possible to invoke the classifier as an external predicate \cite{tplp}.

\subsection{Bringing-in domain knowledge}\label{sec:domK}

The CIP-based specifications we have considered so far allow all kinds of  counterfactual interventions on feature values. However, this may be undesirable or unrealistic in certain applications. For, example, we may not end up producing, and even less, using for score computation, some entities representing people who have the combination of values $\mbox{\sf yes}$ and $\mbox{\sf yes}$ for the propositional features $\mbox{\sf Married}$ and $\mbox{\sf YoungerThan5}$. \ Declarative approaches to specification and computation of counterfactual explanations have the nice feature that {\em domain knowledge and semantic constraints} can be easily integrated with the base specification. Procedural approaches may, most likely,  require changing the underlying code.
We use an example to illustrate the point. For more details and a discussion see \cite{tplp}.

 \begin{example} (example \ref{ex:tree2} continued) It could be that in a particular geographic region, ``raining with a strong wind at the same time" is never possible. \ When producing counterfactual interventions for the entity $\e$, such a combination should not be produced or considered.

 This can be done by imposing a  hard program constraint
{\footnotesize
\begin{verbatim}
 % hard constraint disallowing a particular combination
    :- ent(E,rain,X,strong,tr).
 \end{verbatim} }
 \noindent that we add to the program in Example \ref{ex:tree2}, from which we previously remove the weak constraints we had in \verb+(*)+ (in order not to discard any model for cardinality reasons).
\ If we run  the new program with {\em DLV},  we obtain only the first model in Example \ref{ex:tree2}, corresponding to the counterfactual entity $\e' = \nit{ent}(\mbox{\sf sunny},\mbox{\sf high}, \mbox{\sf weak})$.
 \boxtheorem
 \end{example}

\section{The Generalized $\mbox{\sf Resp}$ Score}\label{sec:resp}

If we want to assign a numerical score to a feature value, say $v=F(\e)$, where $F$ has a relatively large domain, $\nit{Dom}(F)$, it could be the case that counterfactually changing $v$ into $v' \in \nit{Dom}(F)$ changes the label (while leaving the other feature values fixed). However, it could be that for nearly all the other values in $\nit{Dom}(F) \smallsetminus \{v,v'\}$, the label does not change. In this case, we might consider that maybe $v$ is not such a strong reason for the originally obtained label, despite the fact that $v$ is still a counterfactual explanation (with empty contingency set) according to Section \ref{sec:xresp}.

For this reason, it might be better to consider all the possible alternative values for $F$, and define and compute the score in terms of an average of the label values, or an expected value for the label in case we have an underlying probability distribution $P$ on the entity population $\mc{E}$. Such a general version of the $\mbox{\sf x-Resp}$ score was introduced and investigated in \cite{deem}. We briefly describe it starting with the simpler case of counterfactual explanations, i.e. without considering contingency sets. Next, we further generalize the score to consider the latter.  So, in the following, the features do not have to be binary.

Assume that  entity $\e$ has gone through a classifier and we have obtained label $1$, which we   would like to explain. Then, for a  feature $F^\star \in \mc{F}$, we may consider as a score:
\begin{equation}
\mbox{\sf Counter}(\mathbf{e},F^\star) := L(\mathbf{e}) - \mathbb{E}(L(\mathbf{e'})~|~\mathbf{e'}_{\!\!_{\mc{F}\smallsetminus\{F^\star\}}} = \mathbf{e}_{_{\mc{F}\smallsetminus\{F^\star\}}}). \label{eq:count}
\end{equation}
Here, $\e_{_S}$, for $S \subseteq \mc{F}$ is the entity $\e$ restricted to the features in $S$. This score measures the expected difference between the label for $\e$ and those for entities that coincide in feature values everywhere with $\e$ but on feature $F^\star$. Notice the essential counterfactual nature of this score, which is reflected in all the possible hypothetical changes of values for $F^\star$ in $\e$.

A problem with $\mbox{\sf Counter}$ is that changing a single value, no matter how, may not switch the original label, in which case
no explanations are obtained. In order to address this problem, we can bring in {\em contingency sets} of feature values, which leads to the
$\mbox{\sf Resp}$ score introduced in \cite{deem}.

Again, consider $\e \in \mc{E}$, an  entity under classification, for which $L(\e) =1$, and a feature $F^\star \in \mc{F}$. Assume we have:
\begin{enumerate}
\item $\Gamma \ \subseteq \ \mc{F} \smallsetminus \{F^\star\}$, \ a set of features that may end up accompanying feature $F^\star$.
\item $\bar{w} = (w_F)_{F \in \Gamma}$,  \ $w_F \in \nit{Dom}(F)$, \  $w_F \neq \e_F$, i.e.  new values for features in $\Gamma$.
\item $\e' := \e[\Gamma := \bar{w}]$, i.e. reset $\e$'s values for $\Gamma$ as in $\bar{w}$.
\item $L(\e') = L(\e) = 1$,  i.e. there is no label change with $\bar{w}$ (but maybe with an extra change for $F^\star$, in next item).
\item There is $v \in \nit{Dom}(F^\star)$, with \ $v \neq F^\star(\e)$ and $\e'' := \e[\Gamma := \bar{w},F^\star:=v]$.
\end{enumerate}
As in Section \ref{sec:xresp}, \ if \ $L(\e)  \neq  L(\e'') = 0$, \ $F^\star(\e)$ is an {\em actual causal explanation} for $L(\e) =1$, with ``contingency set"  $\langle \Gamma, \e_\Gamma\rangle$, where $\e_\Gamma$ is the projection of $\e$ on $\Gamma$.

In order to define the ``local" responsibility score, make $v$ vary randomly under conditions 1.-5.:
\begin{equation}
\mbox{\sf Resp}(\e,F^\star,\Gamma,\bar{w}) := \frac{L(\e') - \mathbb{E}[L(\e'')~|~\e''_{\mc{F}\smallsetminus \{F^\star\}} = \e'_{\mc{F}\smallsetminus \{F^\star\}}]}{1 + |\Gamma|}. \label{eq:local}
\end{equation}
If, as so far,  label $1$ is what has to be explained, then $L(\e')=1$, and the numerator is a number between $0$ and $1$.  Here, $\Gamma$ is fixed.  Now, we can minimize its size, obtaining the (generalized) responsibility score as the maximum local value; everything relative to  distribution $P$:
\begin{eqnarray}
\mbox{\sf Resp}_{\e,F^\star}(F^\star(\e)) \ &:=& \ \mbox{\large \nit{max}} \ \ \mbox{\sf Resp}(\e,F^\star,\Gamma,\bar{w}) \label{eq:global}\\
&&\mbox{\phantom{oo}}|\Gamma| \ \mbox{\nit{min.}},  \ \mbox{(\ref{eq:local})} >0 \nonumber \\ &&\mbox{\phantom{oo}}{\langle \Gamma, \bar{w}\rangle \models 1.\!-\!4.}\nonumber
\end{eqnarray}
This score was introduced in \cite{deem},  where experiments and comparisons with other scores, namely $\mbox{\sf Shap}$ (c.f. Section \ref{sec:shap}) and the {\em FICO} score \cite{CetA18},  are shown. Furthermore, different probability distributions are considered.
Notice that, in order to compute this score,  there is no need to access the internals of the classification model.

\section{The $\mbox{\sf Shap}$ Score}\label{sec:shap}

In  the context of classification, the Shapley value (c.f. Section \ref{sec:shapy}) has taken the form of the $\shap$ score \cite{LetA20}, which we briefly introduce. Given the binary classifier, $\mc{C}$, on binary entities, it becomes crucial to identify a suitable game function. In this case, it will be expressed in terms of expected values (not unlike the causal-effect score), which requires an underlying probability space on the population of entities, $\mc{E}$.  We will consider, to fix ideas, the {\em uniform probability
  space} on $\mc{E}$. Since we will consider only binary feature values, taking values $0$ or $1$, this is the uniform distribution on  $\mc{E} = \{0,1\}^n$, assigning probability $P^u(\e) = \frac{1}{2^n}$ to $\e \in \mc{E}$. One could consider other distributions \cite{deem,aaai21}.

Given a set of features $\mc{F} = \{F_1, \ldots, F_n\}$, and an entity ${\mathbf{e}}$ whose label is to be explained,  the set of players $D$ in the game is $\mc{F}(\e) :=  \{F(\mathbf{e})~|~  F \in \mc{F}\}$, i.e. the set of feature values of  $\mathbf{e}$. Equivalently, if $\e = \langle x_1, \ldots, x_n\rangle$, then $x_i = F_i(\e)$. We assume these values have implicit feature identifiers, so that duplicates do not collapse, i.e. $|\mc{F}(\e)| = n$.  The game function is defined as follows. \ For $S \subseteq \mc{F}(\e)$,

\vspace{2mm}
\centerline{${\mc{G}_\mathbf{e}(S) := \mathbb{E}(L(\mathbf{e'})~|~\mathbf{e'}_{\!S} = \mathbf{e}_S)}$,}

\vspace{2mm} \noindent where $\mathbf{e}_S$: is the projection of $\e$ on $S$. This is the expected value of the label for entities $\e'$ when their feature values  are fixed and equal to those in $S$ for $\e$. Other than that, the feature values of $\e'$ may independently vary over $\{0,1\}$.

Now, one can instantiate the general expression for the Shapley value in (\ref{eq:sh}), using this particular game function, as ${\nit{Shapley}(\mc{F}(\e),\mc{G}_\mathbf{e},F(\mathbf{e}))}$,
obtaining, for a particular feature value $F(\e)$:
\begin{eqnarray*}
\shap(\mc{F}(\e),\mc{G}_\mathbf{e},F(\mathbf{e})) &:=& \sum_{S\subseteq
  \mc{F}(\e) \setminus \{F(\mathbf{e})\}} \frac{|S|! (n-|S|-1)!}{n!} \times \\
&&\hspace{-0.5cm}(\mathbb{E}(L(\mathbf{e}'|\mathbf{e}'_{S\cup \{F(\mathbf{e})\}} = \mathbf{e}_{S\cup \{F(\mathbf{e})\}})\ - \ \mathbb{E}(L(\mathbf{e}')|\mathbf{e}'_S = \mathbf{e}_S)).
\end{eqnarray*}
Here, the label ${L}$ acts as a Bernoulli random variable that takes values through the classifier.
\ We can see that the $\mbox{\sf Shap}$ score is a weighted average
of differences of expected values of the labels \cite{LetA20}.
\ We may notice that counterfactual versions of the initial entity are implicitly considered.

The $\mbox{\sf Shap}$ score can be applied with black-box classifiers. Under those circumstances its computation takes exponential time in that all permutations of subsets of features are involved. However, sometimes, when the classifier is explicitly available, the computation cost  can be brought down, even to polynomial time. This is the case for several classes of Boolean circuits that can be used as classifiers, and in particular, for decision trees  \cite{LetA20,aaai21,guy}. For other explicit Boolean circuit-based classifiers, the computation of $\mbox{\sf Shap}$ is still $\#P$-hard \cite{aaai21,guy}.

\section{Final Remarks}\label{sec:last}

 Explainable data management and explainable AI  (XAI) are effervescent areas of research.
 The relevance of explanations can only grow, as observed from- and due to the legislation and regulations that are being produced and enforced in relation to explainability, transparency and fairness of data management and AI/ML systems.

There are different approaches and methodologies in relation to explanations, with causality, counterfactuals and scores being prominent approaches that have a relevant role to play.
Much research is still needed on the use of {\em contextual, semantic and domain knowledge}. Some approaches may be more appropriate in this direction, and we argue that declarative, logic-based specifications can be successfully exploited \cite{tplp}.

 Still fundamental research is needed in relation to the notions of {\em explanation} and {\em interpretation}. An always present question is: {\em What is a good explanation?}. \ This is not a new question, and in AI (and other areas and disciplines) it has been investigated. In particular in AI,  areas such as {\em diagnosis} and  {\em causality} have much to contribute.

Now, in relation to {\em explanations scores}, there is still a question to be answered: \ {\em What are the desired properties of an explanation score?}. The question makes a lot of sense, and may not be beyond an answer. After all, the  general
Shapley value emerged from a list of {\em desiderata} in relation to coalition games, as the only measure that satisfies certain explicit properties \cite{S53,R88}. Although the Shapley value is being used in XAI, in particular in its \shap \ incarnation, there could be a different and specific  set of desired properties of explanation scores that could lead to a still undiscovered explanation score.

 \vspace{3mm} \noindent {\bf Acknowledgments: } \ L. Bertossi has been a member of the Academic Network of RelationalAI Inc.,
 where his interest in explanations in ML started. \ Part of this work was funded by ANID - Millennium Science Initiative Program - Code ICN17002. Help from  Jessica Zangari and  Mario Alviano with information about  {\em DLV2}, and from Gabriela Reyes with the {\em DLV} program runs is much appreciated. We are grateful to an anonymous reviewer for valuable comments.

\end{document}